\documentclass{article}

\usepackage[preprint]{neurips_2020}

\usepackage[utf8]{inputenc} % allow utf-8 input
\usepackage[T1]{fontenc}    % use 8-bit T1 fonts
\usepackage{hyperref}       % hyperlinks
\usepackage{url}            % simple URL typesetting
\usepackage{booktabs}       % professional-quality tables
\usepackage{amsfonts}       % blackboard math symbols
\usepackage{nicefrac}       % compact symbols for 1/2, etc.
\usepackage{microtype}      % microtypography

\usepackage{multicol}
\usepackage{subfigure}
\usepackage{graphicx}
\usepackage{amsmath}
\usepackage{multirow}
\usepackage{verbatim}
\usepackage{amssymb}
\usepackage{algorithm}
\usepackage{algpseudocode}
\usepackage{algorithmicx}
\usepackage{caption}
\usepackage{setspace}
\usepackage{appendix}

\title{Adding A Filter Based on The Discriminator to Improve Unconditional Text Generation}

\author{Xingyuan Chen$^{1,3}$
\ Ping Cai$^2$
\ Peng Jin$^{3,}$\thanks{Peng Jin, Xinyu Dai and Jiajun Chen are the co-corresponding authors.}
\ \;Hongjun Wang$^2$
\ Xinyu Dai$^1$
\ Jiajun Chen$^1$  \\
\
\\$^1$ Department of Computer Science and Technology\\
Nanjing University, Nanjing, China \\
\texttt{1045258214@qq.com;daixinyu@nju.edu.cn;chenjj@nju.edu.cn} \\
\
$^2$School of Information Science and technology\\
Southwest Jiaotong University, Chengdu, China \\
\texttt{1061185275@qq.com;wanghongjun@swjtu.edu.cn} \\
\
$^3$School of Computer Science\\
Leshan Normal University, Leshan, China \\
\texttt{jandp@pku.edu.cn} \\
}

\begin{document}

\maketitle

\begin{abstract}
  The autoregressive language model (ALM) trained with maximum likelihood estimation (MLE) is widely used in unconditional text generation. Due to exposure bias, the generated texts still suffer from low quality and diversity. This presents statistically as a discrepancy between the real text and generated text. Some research shows a discriminator can detect this discrepancy. Because the discriminator can encode more information than the generator, discriminator has the potentiality to improve generator. To alleviate the exposure bias, generative adversarial networks (GAN) use the discriminator to update the generator's parameters directly, but they fail by being evaluated precisely. A critical reason for the failure is the difference  between the discriminator input and the ALM input. We propose a novel mechanism by adding a filter which has the same input as the discriminator. First, discriminator detects the discrepancy signals and passes to filter directly (or by learning). Then, we use the filter to reject some generated samples with a sampling-based method. Thus, the original generative distribution is revised to reduce the discrepancy. Two ALMs, RNN-based and Transformer-based, are experimented.  Evaluated precisely by three metrics, our mechanism consistently outperforms the ALMs and all kinds of GANs across two benchmark data sets.
\end{abstract}

\section{Introduction}
Text generation is an important part of many applications such as machine translation \cite{bahdanau2014neural,Wu2016Google}, dialog systems \cite{Du2019,Ghosh2017affect}  and image caption generation \cite{Liu2017Auto,Vinyals2015show}. Unconditional text generation, which generates novel, reasonable and meaningful sentences, is a stepping stone for the above tasks, thus becoming a hot topic recently \cite{Yu2016SeqGAN,Fedus2018MaskGAN,dAutu2019Scratch}.

Trained with MLE, an ALM \cite{mikolov2010recurrent} which is a Long Short-Term Memory (LSTM) \cite{hochreiter1997long} or Transformer \cite{Vaswani2017Attention} architecture, has been widely used as text generators \cite{Graves2013rnn,Radford18}. However, due to some limitations including exposure bias \cite{Bengio2015Scheduled}, generated texts still suffer from low quality in regard to semantics and global coherence and are not even perfect grammatically speaking \cite{Caccia2019falling}.

In fact, both low quality and poor diversity present statistically as a large discrepancy between the distribution of real text and the distribution of the generated text. To improve an unconditional text generator, we have to not only improve sample quality but also make the kinds and ratios of modes approximate the real texts as much as possible. This is equivalent to reducing the distributional discrepancy.

%Many researchers improve the quality at the cost of degrading the diversity.
The discriminator can detect this discrepancy because it encodes more tokens (even the entire sequence) than ALM. For example, at the time step $t$, the generator predicts $x_t$ given $[x1,...,x_{t-1}]$, but, the discriminator predicts true or false given $[x1,...,x_{t-1}, x_t]$ (even the entire sequence). This power of discriminator has been used to improve several text generation tasks such as text abstraction \cite{Scialom2020Abstractive}, fake news detection \cite{Zellers2019} and pre-training ALM \cite{Clark2019Electra}. It should be noted that we use CNN instead of RNN as the discriminator, and its input is the entire sequence.

%In this paper, we use this potentiality of discriminator to improve unconditional text generation by adding a filter after the generator.

GAN \cite{Goodfellow2014Generative} has been advocated as a means to improve a generator by many researchers\cite{Yu2016SeqGAN,Fedus2018MaskGAN,Nie2019ICLR}. In this way, an ALM is pre-trained with real text \footnote{An exception is RelGAN.} and then being used as a generator. A discriminator is trained with generated texts and real texts. The parameters of the generator are updated according to the detected discrepancy signal from the discriminator. However, \cite{Caccia2019falling,Semeniuta2018On} argue GANs sacrifice diversity to improve quality. Evaluated by temperature sweep, \cite{Caccia2019falling} finds ALM beats all GANs in a quality-diversity space. A critical reason for the failure is the difference  between the discriminator input and the ALM input. This difference causes gradient estimation and optimization instability.

%Inspired by GAN, the discriminator can detect this discrepancy between generated text and real text\cite{Zellers2019}.
To avoid the problems caused by different inputs, we introduce a filter, which has the same input as the discriminator. First, a discriminator is well trained as a binary text classifier with the full sentences from both the real and the generated samples. Then, a filter is added after the generator, which depends on the well-trained discriminator, to reject some generated texts. Thus, a new generator is created by this integration. The accepted samples are the output of this new generator. The discrepancy of the new generator is smaller than that of the old one. There are several ways to implement the filter based on discriminator. In this paper, a sampling-based filtering method is implemented. Meanwhile, an generative adversarial based method is suggested.

All code and data are available at GitHub\footnote{\url{https://github.com/anonymous1100/D_Improves_G_without_Updating_G}}. Our contributions are listed as follows:

\begin{itemize}
\item We demonstrate the discriminator can reliably improve an autoregressive text generator, by combining a filter to this generator to create a new generator.
\item We implement a sampling-based filtering method. A generative adversarial based method is proposed further. %This method provide a new way to balance quality and diversity.   %
\item Evaluated by three metrics, our method consistently improves both RNN-based and Transformer-based ALMs across two benchmark data sets.

\end{itemize}

\section{The Filtering Mechanism}
\subsection{ALM Trained with MLE as the Text Generator}
To model the distribution of real text sequences, which is denoted as $p_r(x)$, $x=[\mathrm{x_1},\mathrm{x_2}...,\mathrm{x_T}]$, an autoregressive neural language model $G_\theta(x)$ is defined as follows,
\begin{equation}
    \label{E1:LM}
    p_\theta(x) = \prod_{t=1}^{T} p_\theta(\mathrm{x_t}|\mathrm{x_1},...,\mathrm{x_{t-1}})
\end{equation}
%$p_\theta(x)$ is the distribution of the sentences which are generated by  $G_\theta(x)$.
$G_\theta(x)$ is usually trained with maximum likelihood estimation,
\begin{equation}
    \label{E2:MLE}
    \mathop{\arg\max}_{\theta}E_{x\sim{p_r(x)}}logp_\theta(x)
\end{equation}

\begin{figure}[ht]
\centering
\subfigure[GAN]{%ye yinggai fencheng laingfu tu!!!
\includegraphics[width=0.4\columnwidth]{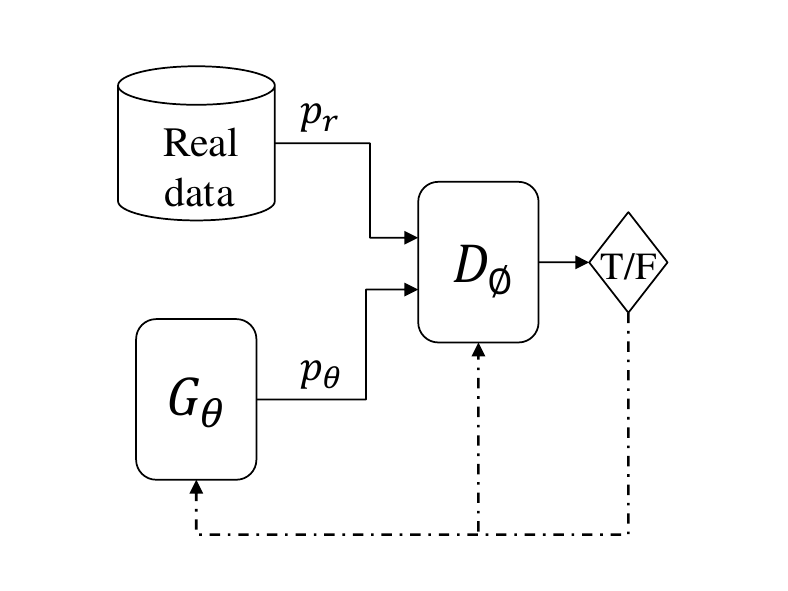}
}
\quad
\subfigure[Filtering Mechanism]{
\includegraphics[width=0.4\columnwidth]{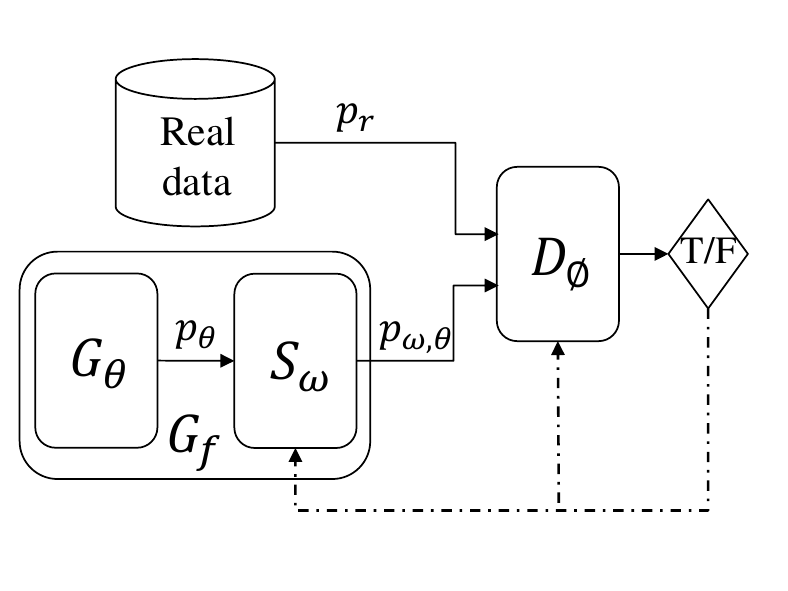}
}
\caption{Our filtering mechanism versus GAN. A filter $S_\omega(x)$ is added to determine the acceptance probability of a sample which is generated by $G_\theta$. This acceptance probability is computed according to feedback signals from discriminator $D_\phi$.} %Filter co-operate with discriminator to revise the distribution of generated samples.}
\label{p1}
\end{figure}

There are two drawbacks of the autoregressive generative model. One is the generation error caused by the exposure bias \cite{Bengio2015Scheduled}, and this error becomes larger as the sequence becomes longer. The other is the uni-directional input brought by the autoregressive method. When the $t$-th token $\mathrm{x_t}$ is generated, the autoregressive generator only encodes sub-sequence [$\mathrm{x_1}$,...,$\mathrm{x_{t-1}}$], but it can't see $\mathrm{x_t}$. On the contrary, the discriminator can encode subsequence [$\mathrm{x_1}$,...,$\mathrm{x_{t-1}}, \mathrm{x_t}$], even the entire sequence, if it is based CNN\cite{Kim2014CNN} or Transformer\cite{Vaswani2017Attention}. It means that the discriminator has the potentiality to improve the generator. In this paper, the discriminator is a CNN and the input is the entire sentence.

%a discriminator $D_\phi$ can encode the whole sequence thus can exploit more context information than the autoregressive model,
%A discriminator can detect the discrepancy between the two distributions of real text and generated text. Let $p_r(x)$ denote the distribution of real data. For the purpose of generality, let $p_g(x)$ denote the distribution of generated text which is generated by any generator. Specifically, for the above mentioned $p_\theta(x)$, there will be $p_g(x) = p_\theta(x)$.

\subsection{The Discriminator can Detect the Discrepancy}
According to \cite{Yu2016SeqGAN,Nie2019ICLR}, a discriminator $D_\phi$  can detect the discrepancy between real sentences and generated ones which are generated by $G_\theta$. To detect this discrepancy, $D_\phi$ needs to be optimized with the following objective function:
\begin{equation}
    \label{E3:optD}
    \mathop{\max}_{D_\phi}(V(D,G)) = \mathbb{E}_{x\sim{p_r(x)}}[logD_\phi(x)] +
    \mathbb{E}_{x\sim{p_\theta(x)}}[log(1-D_\phi(x))]
\end{equation}
Assuming $D_\phi^{*}(x)$ is the optimal resolution of the above function, according to \cite{Goodfellow2014Generative}, it will be,
\begin{equation}
    \label{E4:Dstar}
    D_\phi^{*}(x) = \frac{p_r(x)}{p_r(x) + p_\theta(x)}
\end{equation}
It is hard to obtain this optimal discriminator. The researchers estimate it with some training methods and demonstrate these approximations of $D_\phi^{*}$ detect the discrepancy effectively \cite{Goodfellow2014Generative,Clark2019Electra}.

\subsection{Adding A Filter to Change Distribution of Generated Samples}
Given a sample $x$ which is generated by $G_\theta$, $D_\phi^{*}(x)$ evaluates the discrepancy between $p_r(x)$ and $p_\theta(x)$. For $x$, we can accept this sample in a certain probability (we call this acceptance probability) by using sampling. For this, we introduce a filter $F_\omega$, which is defined as a filter function,
\begin{equation}
    \label{E51:Somega}
    s = S_\omega(x), 0\leq s \leq 1
\end{equation}
which is used to determine the acceptance probability of a generated sample. Therefore, we obtain a new distributional function as follows.
\begin{equation}
    \label{E51:pOmega}
    p_{\omega,\theta}(x) = \frac{1}{c}S_\omega(x)p_\theta(x)
\end{equation}
$\frac{1}{c}$ is a normalization factor and $c=\int S_\omega(x)p_\theta(x)dx$. Obviously, $c$ is the acceptance ratio, i.e. the number of accepted samples divides the number of all generated samples. Thus, a new generator $G_{f}$ is created by binding $F_\omega$ after $G_\theta$.  $p_{\omega,\theta}(x)$ is its generative distribution. Figure~\ref{p1} illustrates this filtering mechanism.
%We try to reduce the gap by filtering which is illustrated in the right of Figure~\ref{p1}(a). This novel mechanism does not change the parameters of $G_\theta$ directly, but creates a new generator $G_{f}$ by binding a  $D_\phi$ after $G_\theta$.  We describe this process in detail.

%According to equation \ref{E4:Dstar}, when $p_\theta(x)$ deviates away from $p_r(x)$ further, $D_\phi^{p_\theta}(x)$ is smaller, thus, $x$ should be removed.

In practice, there are many methods to design and train a filter according to the detected discrepancy signals from $D_\phi(x)$.  In next section, we implement a simple but powerful sampling-based method in which a LSTM based or Transformer based ALM is used as the generator, and CNN is used as the discriminator. It is important to note that in this framework, the parameters of filter are updated rather than that of the generator. Meanwhile, the filter and the discriminator have the same input. Therefore, various problems caused by the difference of the inputs of the generator and the discriminator in the original GANs are avoided. Further, A general filter formulation is suggested.

\subsection{Implement the Filter by Sampling}
%How to filter generated text is a challenge. In this section, we describe a threshold-based method and analyze it theoretically. Finally, the threshold-based method can be used recursively for improving performance further.

To reduce the distribution discrepancy as much as possible with the signal detected by the discriminator, we make $p_{\omega,\theta}(x)$ of $G_f$ equal to the real distribution $p_{r}(x)$ when a generated sample $x$ needs to be filtered. Therefore, $S_\omega(x)$ is obtained,
\begin{equation}
    \label{E5x:Smapling}
    S_\omega(x) =
    \begin{cases}
        \frac{cD_\phi(x)}{1-D_\phi(x)} \quad D_\phi(x) < u_c \\
       1 \qquad \qquad D_\phi(x) \geq u_c \\
    \end{cases}
\end{equation}
$c$ is the acceptance ratio in equation \ref{E51:pOmega}. $u_c$ is the sampling boundary, and its value is determined by $c$. Generally, the smaller $c$ is, the larger $u_c$ is. If $D_\phi(x) = D_\phi^{*}(x)$, there will be  $p_{\omega,\theta}(x) = p_r(x)$ when $D_\phi(x) < u_c$.

Because we can only obtain the approximation of $D_\phi^{*}(x)$, i.e. $D_\phi(x) \neq D_\phi^{*}(x)$, there is still a discrepancy between $p_{\omega,\theta}(x)$ and $p_r(x)$. Thus, we could train another discriminator to detect the new discrepancy. Therefore, the filtering method can be iterated. To avoid adding additional filters, a general filtering mechanism is suggested as follows,
\begin{equation}
    \label{E52:optDOmega}
    \mathop{\min}_{S_\omega}\mathop{\max}_{D_\phi}(V(D,G)) = \mathbb{E}_{x\sim{p_r(x)}}[logD_\phi(x)] + \\
    \mathbb{E}_{x\sim{p_{\omega,\theta}(x)}}[log(1-D_\phi(x))]
\end{equation}
where $S_\omega$ is a learnable function (a neural network). Within this GAN framework, both the discriminator $D_\phi$ and the filter $S_\omega$ are updated in adversarial learning. This method has the potential to improve the performance further. The training of this framework is an open problem.

%How to train $S_\omega$ is an open problem.
%这个被建议的方法有潜力进一步提升基于过滤的文本生成。

\section{The Estimation of Optimal Discriminator and Sample Boundary}
\subsection{The Estimation of the Optimal Discriminator}
A challenge is that we can not obtain the optimal function $D_\phi^{*}(x)$ directly.  To find an approximated function, we follow \cite{Zhu2018Texygen} to use a CNN-based neural network as the discriminator $D_\phi$, because CNN is very powerful for text classification \cite{Kim2014CNN,Lai2015tc}. Supervised learning is applied to minimize cross entropy. In order to avoid imbalanced learning \cite{He2009Imbalaced}, we always generate as many sentences as we have real sentences. According to equation~\ref{E3:optD}, $D_\phi$ is trained until it converges. This convergent discriminator is denoted as $\hat{D}_\phi^{*}(x)$ which is the approximated function of $D_\phi^{*}(x)$.

This training process needs many epochs to make the discriminator convergence. This training strategy is different completely from GAN, which only trains the discriminator tens of batchs in each round adversarial learning in order to avoid the local optima.

For comparison with language GANs, the architecture and hyper-parameters of our discriminator are the same as the discriminator used in SeqGAN \cite{Yu2016SeqGAN}.

\subsection{The Estimation of Sample Boundary}
The value of the sampling boundary $u_c$ is determined by the acceptance ratio $c$. According to formula \ref{E5x:Smapling}, $u_c$ has a unique value corresponding to $c$, because $u_c$ becomes small as $c$ increase. The detail implement is described by algorithm~\ref{al:alg1}.
%因为u_c随着c增大而变小，于是c给定后，u_c有唯一的值和c对应。
\begin{table}[H]
    \begin{minipage}{0.48\columnwidth}
        \begin{algorithm}[H]
            \centering
            \footnotesize
            \captionof{algorithm}{: The estimation of $u_c$}
            \label{al:alg1}
            %\footnotesize{\caption{: The estimation of $u_c$}} %according to $D_\phi(x)$}
            \begin{algorithmic}[1]
                \Require \\
                    %Initializing $u_c$ in $0\sim1$ randomly\\
                    $u_c=0.5,ds=0.01$, $\mathcal{M}=1000$, $\mathcal{N}=100$
                    %A threshold loss value, $L_{th}$; Number of training, $\mathcal{N}$\\
                    %A random start position in NE list for replacement, $s$
                %\Ensure
                    %A convergenced $c$
                %\State $i \gets 1$, $j \gets 0$
                \For{$s = 1 : \mathcal{N}$}
                    \State sum\_acp = 0
                    \For{$k = 1 : \mathcal{M}$}
                        \State generate $x\sim{G_\theta(x)}$, sample $z\sim{U[0,1]}$
                        \If{$D_\phi(x)>u_c$ or $z>S_\omega(x)$}
                            \State $sum\_acp \gets sum\_acp + 1$
                        \EndIf
                    \EndFor
                    \State $c\_1 \gets sum\_acp/\mathcal{M}$ \% get acceptance ratio
                    \If{c\_1 < c} \qquad \quad \% update $u_c$
                        \State $u_c \gets u_c - ds$
                    \Else
                        \State $u_c \gets u_c +  ds$
                    \EndIf
                \EndFor
            %\State \Return{$c$}
                %\State
            \end{algorithmic}
        \end{algorithm}
    \end{minipage}
    \hfill
    \begin{minipage}{0.48\linewidth}
        \centering
        \caption{The values of Hyper-parameters. For two generators, the values of GPT-2 are listed in parentheses when they are different from LSTM.  For discriminator which is a convolutional neural network, the “window size, kernel numbers” of each convolutional layer is listed and the dropout is not adapted. "Pa." denotes the parameter and "lr" is the learning rate.}
        \begin{tabular}{c|c|c|c}
        \toprule[0.8pt]
        %\hline
        \small{\textbf{Pa. of G}} & \small{\textbf{Value}} & \small{\textbf{Pa. of D}} & \small{\textbf{Value}}\\
        \hline
        \small{$d$\_model} & \small{512(768)} & \small{layer1} & \small{2,100}\\
        \small{layer} & \small{2(12)}  & \small{layer2} & \small{3,200}\\
        \small{dropout} & \small{0.5(0.1)}  & \small{dropout} & \small{0.0}\\
        \small{lr} & \small{1e-3(2e-5)} & \small{lr} & \small{1e-4}\\
        \small{head}  & \small{12} & \small{---} & \small{---} \\
        \small{batch size} & \small{128} &   \small{batch size} & \small{512} \\
        \small{\# of pa.}  & \small{30.1(268)M} &   \small{\# of pa.} & \small{10.5M} \\
        %\hline
        \bottomrule[0.8pt]
        \end{tabular}
        \label{t:hyper}
    \end{minipage}
\end{table}

\section{Experiments}
%We experiment on two benchmark data sets with the threshold-based method. Experimental settings, results and analysis are described as follows.

\subsection{Experimental Settings}
\textbf{Data sets.} Two benchmark data sets recently have been used widely for unconditional text generation. One is COCO Image captions \footnote{http://cocodataset.org/} \cite{Chen2015coco} which consists of relatively short sentences and small amounts of sentences. For comparison, we follow \cite{Caccia2019falling} to pre-process this data set. There are in total 4,633 word types and the longest sentence consists of 37 words. Both the training and test data contain 10,000 sentences. The average length of sentences is about 11 words. The other one is EMNLP2017 WMT News \footnote{http://www.statmt.org/wmt17/} which consists of relatively long sentences and large amounts of sentences.
Once again, we follow \cite{Caccia2019falling} to pre-process this data set. As a result, it consists of about 280k sentences and the sentences' average length is about 20 words. There are in total 5,697 word types and the longest sentence consists of 51 words. 10,000 sentences are used as the test data. Among all the training data, the last 10,000 sentences are used as validation data.

\textbf{Evaluation Metric.}
The sample diversity is as important as the sample quality for unconditional text generation. \cite{Zhu2018Texygen} proposes  BLEU \cite{Papineni2002BLEU} versus Self-BLEU, which are most frequently used by the research community. To use this paired metric precisely, \cite{Caccia2019falling} proposes a temperature sweep technique by adjusting the softmax temperature, thereby drawing a curve rather than only plotting a dot in the quality-diversity space. Due to BLEU biases local consistency, \cite{DBLP:journals/corr/abs-1804-07972,Semeniuta2018On} measure global semantic by language model score (measuring sample quality) verse language model score (measuring sample diversity). We follow \cite{Caccia2019falling} to use both local and global metrics with this technique to evaluate the models.

Further, considering BLEU versus self-BLEU as a two-dimension metric, we follow \cite{dAutu2019Scratch} to use a single metric - Fr\'{e}chet Embedding Distance (FED). It computes the Fr\'{e}chet distance between two Gaussian distributions. The same embeddings trained from a Universal Sentence Encoder\footnote{The model is available at https://tfhub.dev/google/universal-sentence-encoder/3} as \cite{dAutu2019Scratch} are adapted. This metric can capture both local and global consistency.

\textbf{Generator and discriminator.}
For our generators, one is LSTM. The other is GPT-2 \cite{Radford19}, which is based on Transformer and achieves the state-of-the-art performance. For LSTM, all the hyper-parameters and the architecture are the same as \cite{Caccia2019falling}. For GPT-2, We optimize some hyper-parameters by observing its performance on validation data. The best settings are adapted. Table~\ref{t:hyper} lists them in detail.

For the discriminator, we follow \cite{Zhu2018Texygen} to use a CNN. Both the hyper-parameters and architecture remain unchanged. To reduce the parameters, all word embeddings are copied from the generator and fixed. We observe the accuracy of classification on validation to make sure the convergence of the discriminator. Because there is no validation data available for COCO Image Captions, we set aside the last 1,000 sentences of the training data as the validation data.

\textbf{Baseline Models.} Two kinds of models are used as baselines. One is ALM which is trained with MLE. Both GPT-2 and LSTM are experimented. The other is language GAN. The performance of SeqGAN and MaliGAN on COCO are taken from \cite{Zhu2018Texygen}. The performance of FMGAN and Rel-GAN are taken from \cite{Caccia2019falling}. All language GANs fine tunes LSTM.

\subsection{Main Results}
Our model consistently outperforms the neural language models trained with MLE and language GANs on three metrics across two benchmark data sets.

\begin{figure}[ht]
\centering
\subfigure[COCO Image Captions]{
\includegraphics[width=0.4\columnwidth]{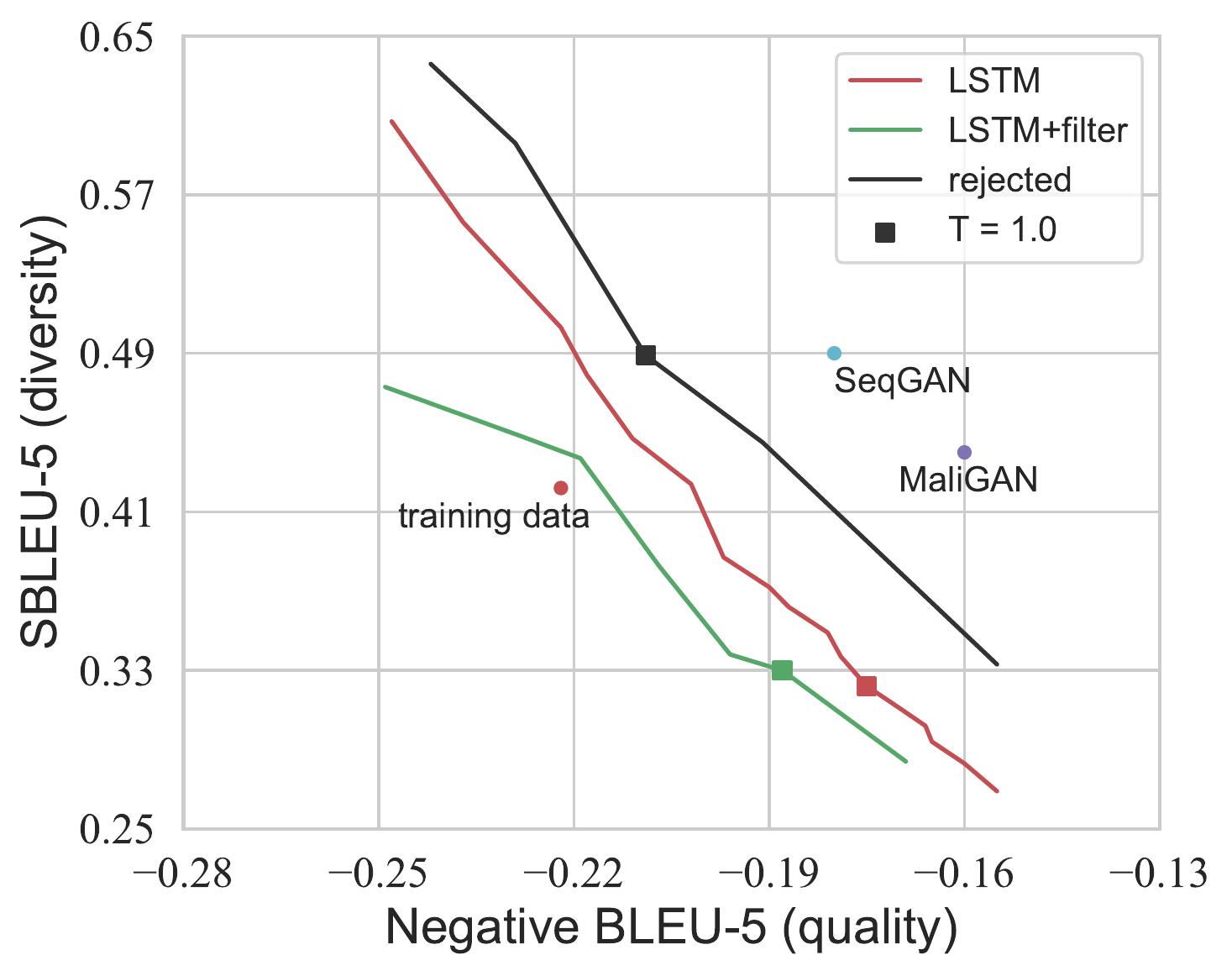}
}
\quad
\subfigure[EMNLP2017 WMT News]{
\includegraphics[width=0.4\columnwidth]{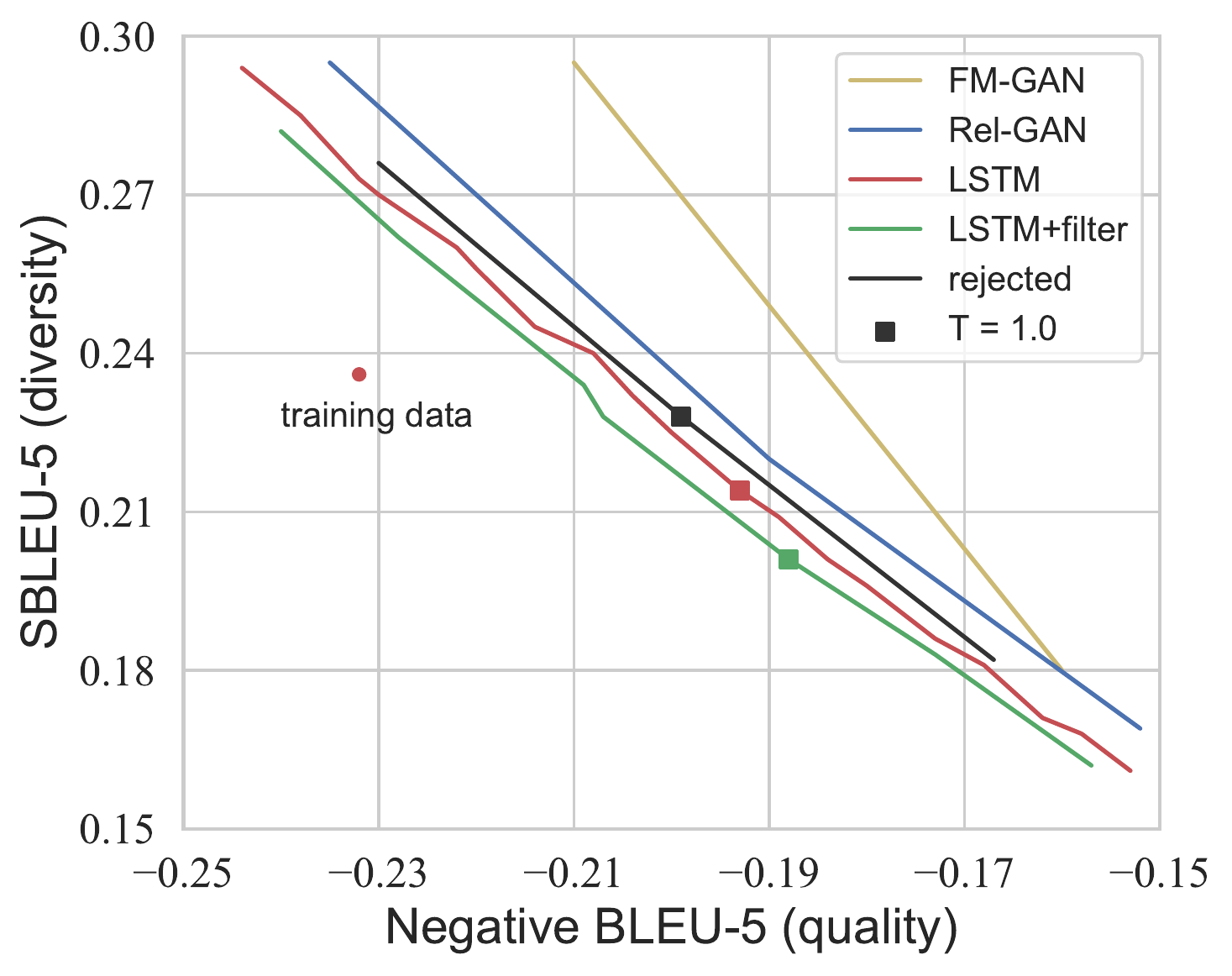}
}
\caption{Local metrics. Lower is better. The results of language GANs on left figure are taken from \cite{Lu2018Past} directly . The results of language GANs on right figure are take from \cite{Caccia2019falling}. $T=1.0$ denotes softmax temperature is 1.0; when $T>1.0$ quality increases but diversity decreases.}
\label{fig:final}
\end{figure}

\begin{figure}[ht]
\centering
\subfigure[LSTM]{
\includegraphics[width=0.4\columnwidth]{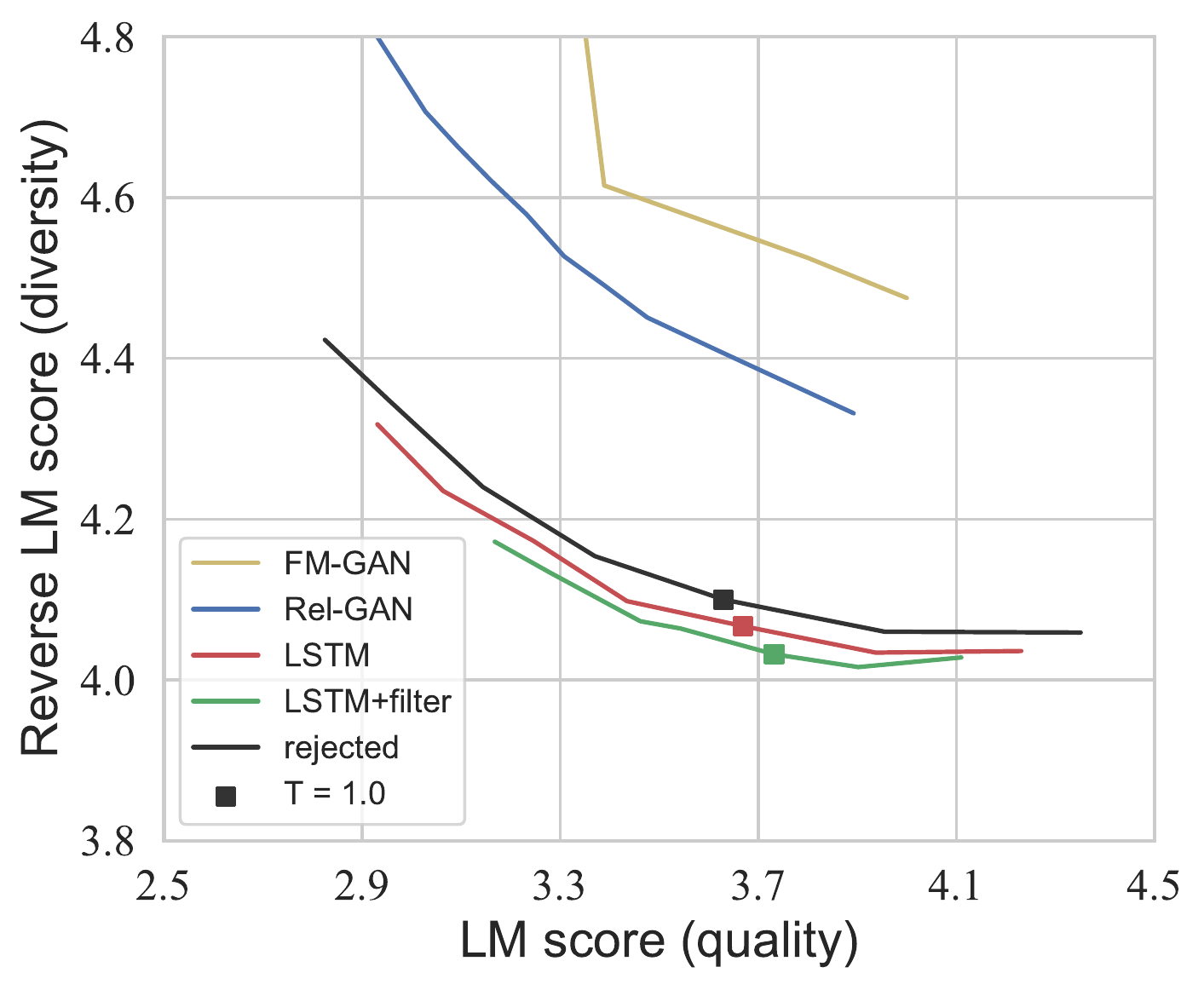}
}
\quad
\subfigure[GPT-2]{
\includegraphics[width=0.4\columnwidth]{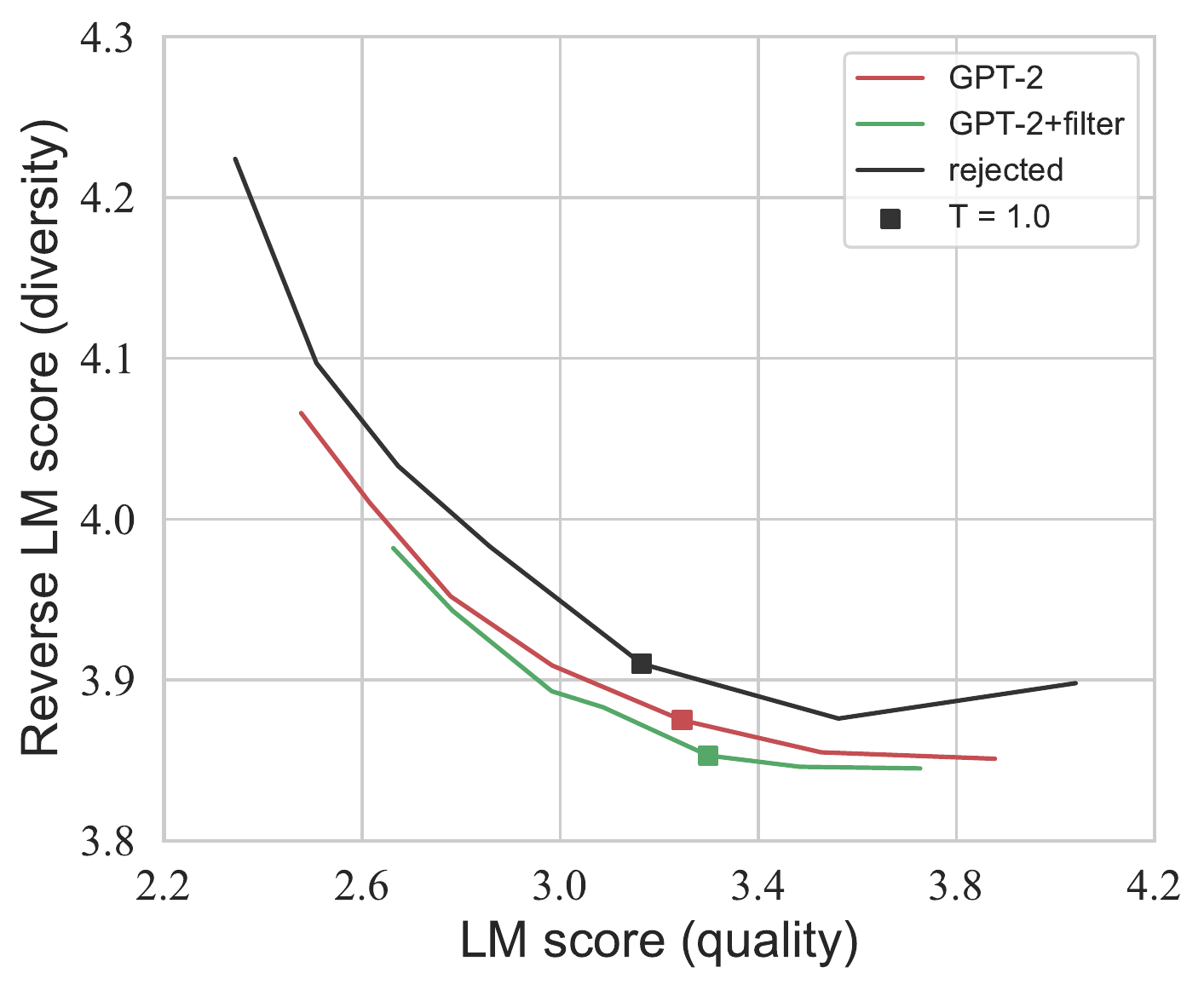}
}
\caption{Global metrics. Lower is better. The results of language GANs on left figure are taken from \cite{Caccia2019falling} directly. The data set is EMNLP WMT News.}
\label{fig:LMRLM}
\end{figure}
For local metric, BLEU versus self-BLEU, we follow \cite{Caccia2019falling} to draw quality-diversity curves at various temperatures (from 0.9 to 1.2). Figure~\ref{fig:final} shows the performance of the new generators always lies at the bottom left of both the red ones, which denote the performance of ALM trained based on MLE. This indicates our filtering-based generators outperform ALM. For comparison, the performance of rejected samples is also plotted as black lines. They lie at the top right of the red ones. This shows the distribution discrepancy between real sentences and the rejected ones are larger. Further, those language GANs are worse than ALM. There is no gain by updating the generator's parameters directly. On the contrary,  the performance is improved by revising the distribution of generated samples with a filter. %A similar situation occurs in GPT-2 and the results are listed in the supplements.

For global metric, language model score versus reverse language model score, we still follow \cite{Caccia2019falling} to report the results with a temperature sweep in figure \ref{fig:LMRLM}.  The filter works well with both LSTM and GPT-2. New generators outperform the old ones and language GANs again. The performance of rejected samples is plotted as black lines and is lower than ALM.  %The results on COCO Image Captions are supplemented.

\begin{figure}[!h]
\centering
\quad
\subfigure[\scriptsize{COCO Image Captions}]{
\includegraphics[width=0.4\columnwidth]{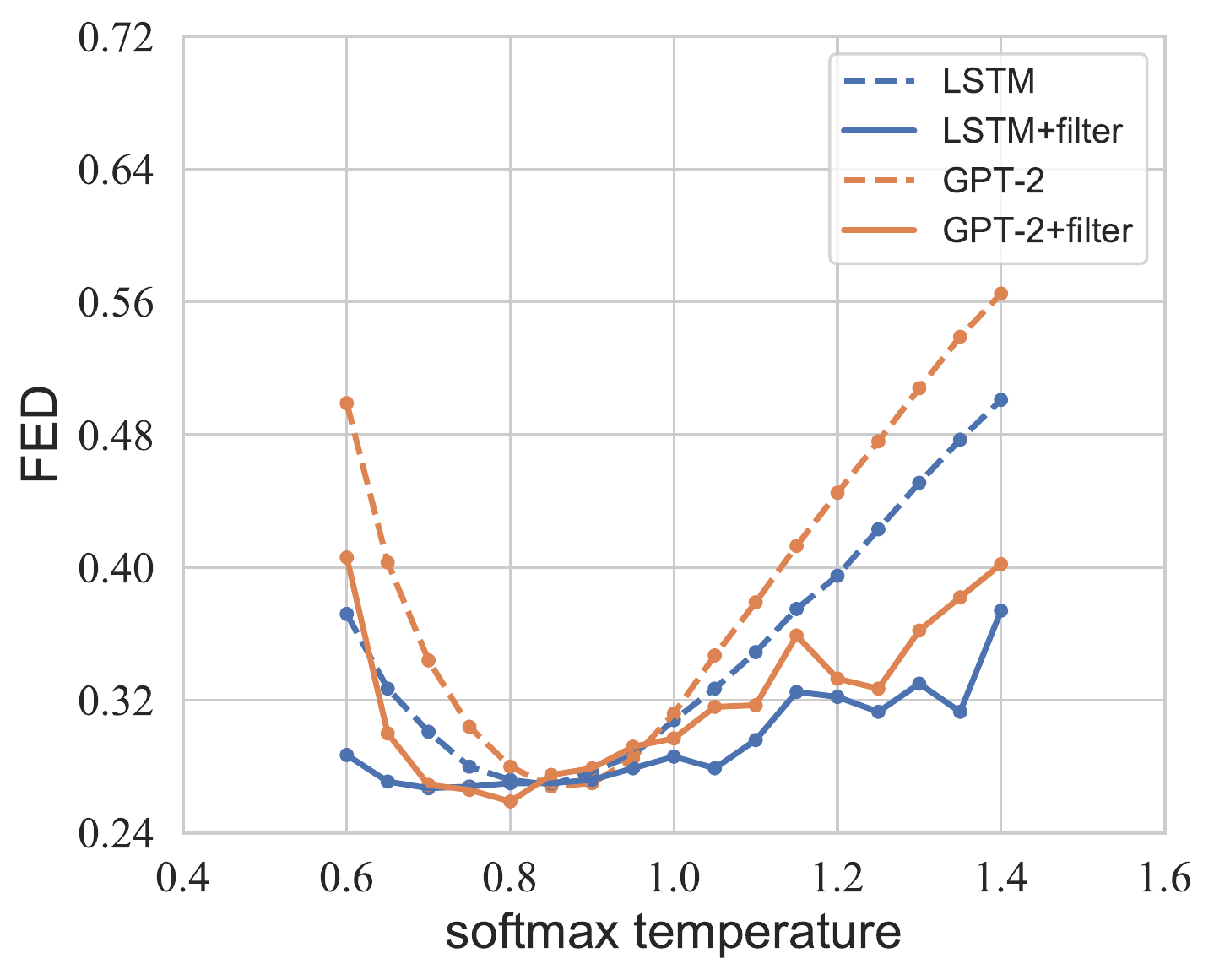}
}
\quad
\subfigure[\scriptsize{EMNLP2017 WMT News}]{
\includegraphics[width=0.4\columnwidth]{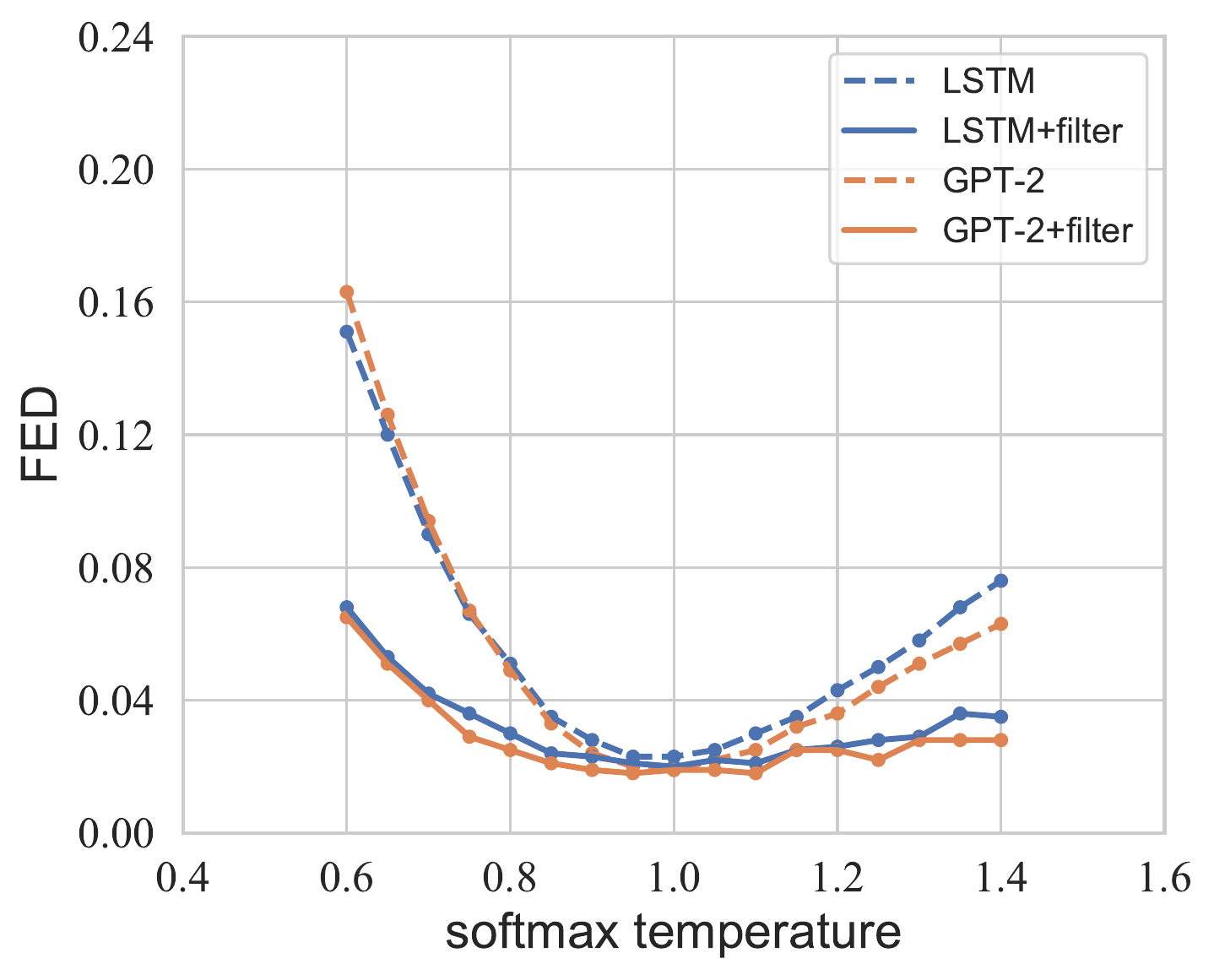}
}

\caption{Single metric. Lower is better.}
\label{fig:FED}
\end{figure}

For single metric, FED, we follow \cite{dAutu2019Scratch}  and find our model outperforms both GPT-2 and LSTM at various softmax temperatures (shown by Figure~\ref{fig:FED}). This demonstrates our method improves ALMs in both local consistency and global semantic. Especially, the further the softmax temperature is from 1.0, the more benefits are obtained.

\section{Analysis}
The acceptance ratio, i.e. the number of the accepted sentences divided by the total generated sentences is efficiency of filtering mechanism. Obviously, the efficiency becomes low as $c$ decreases. We analyze the influence of acceptance ratio in detail.

\subsection{The Influence of Acceptance Ratio}
\begin{figure}[H]
\begin{minipage}{0.48\linewidth}
\centering
\includegraphics[width=0.8\columnwidth]{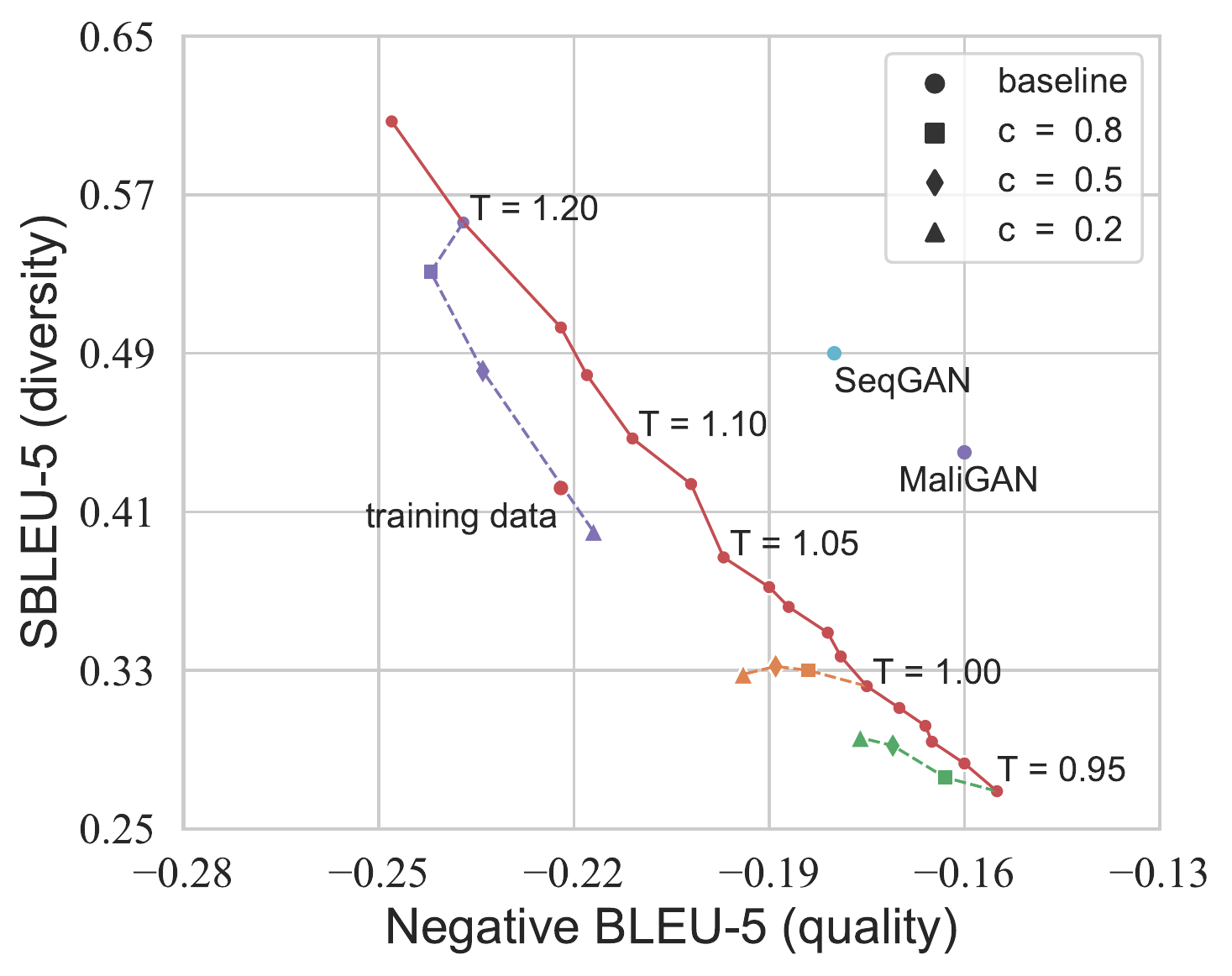}
\caption{The influence of acceptance ratio.}
\label{fig:bsb}
\end{minipage}%
\hfill
\begin{minipage}{0.48\linewidth}
\centering
\captionof{table}{The sample boundary ($u_c$) and the error ratio (difficulty of classification) vary with the different acceptance ratios ($c$) at  temperature 1.0. Baseline is $c=1.0$, i.e. no filter is added. The minimum of $c$ is 0.2 and $u_c=1.0$}%  LSTM is the generator.}
\begin{tabular}{c|cc|cc}
\toprule[0.8pt]
          & \multicolumn{2}{c|}{\scriptsize{sample boundary}} & \multicolumn{2}{c}{\scriptsize{classification error}} \\ \hline
\footnotesize{Data sets} & \scriptsize{COCO}             & \scriptsize{EMNLP}            & \scriptsize{COCO}            & \scriptsize{EMNLP}          \\ \hline
baseline & ---            & ---            & 0.325           & 0.312          \\ %\hline
$c=0.8$ & 0.010            & 0.002            & 0.374           & 0.334          \\ %\hline
$c=0.5$ & 0.160            & 0.018            & 0.400           & 0.360          \\ %\hline
$c=0.2$ & 1.000            & 1.000            & 0.408           & 0.370         \\
\bottomrule[0.8pt]
\end{tabular}
\label{t:accept&accuracy}
\end{minipage}
\end{figure}

The filtering mechanism always works well regardless of the value of $c$. Figure~\ref{fig:bsb} shows the influence of the acceptance ratio with LSTM on the COCO Image Captions data set. We find all points move towards performance point of the training data. Our model even achieves the BLEU score verse self-BLEU score almost equal to scores of the training data at temperature 1.2. This highlights the effectiveness of our method.%This is the limitation of this metric. \cite{dAutu2019Scratch} also illustrates this by using a simple 5-gram statistical language model with Kneser-Ney smoothing.

Acceptance ratio plays a role in balancing quality and diversity, and has a huge impact on them. Thus, this provides us a new way to adjust the sample quality and sample diversity. Acceptance ratio can be used in conjunction with the temperature to obtain better generation performance. % Before that, only what we can do is adjusting the softmax temperature. On the contrary,  the quality is improved greatly and the diversity is sacrificed a little, when the temperature is less than or equal 1.0.

%这个控制变量可以和温度变量配合使用，来获得更好的生成性能。
%The diversity increases dramatically but the quality decreases slightly,  as the acceptance ratio becomes small with the temperature is greater than 1.0.

The left two columns of table \ref{t:accept&accuracy} show the variation of the sample boundary $u_c$ with the acceptance ratio. As the acceptance ratio decreases, the value of $u_c$ increases. Thus, more and more generated samples will be considered to be rejected in a certain probability according to equation \ref{E5x:Smapling}. The right two columns of table \ref{t:accept&accuracy} summary the influence of the acceptance ratio on classification error rate. Once the filter is added, the error rate decrease. Further, the error rate increases with the decrease of $c$.

\subsection{The Difficulty of Classification}
\begin{figure}
\begin{minipage}[t]{0.48\linewidth}
\centering
\includegraphics[width=0.8\columnwidth]{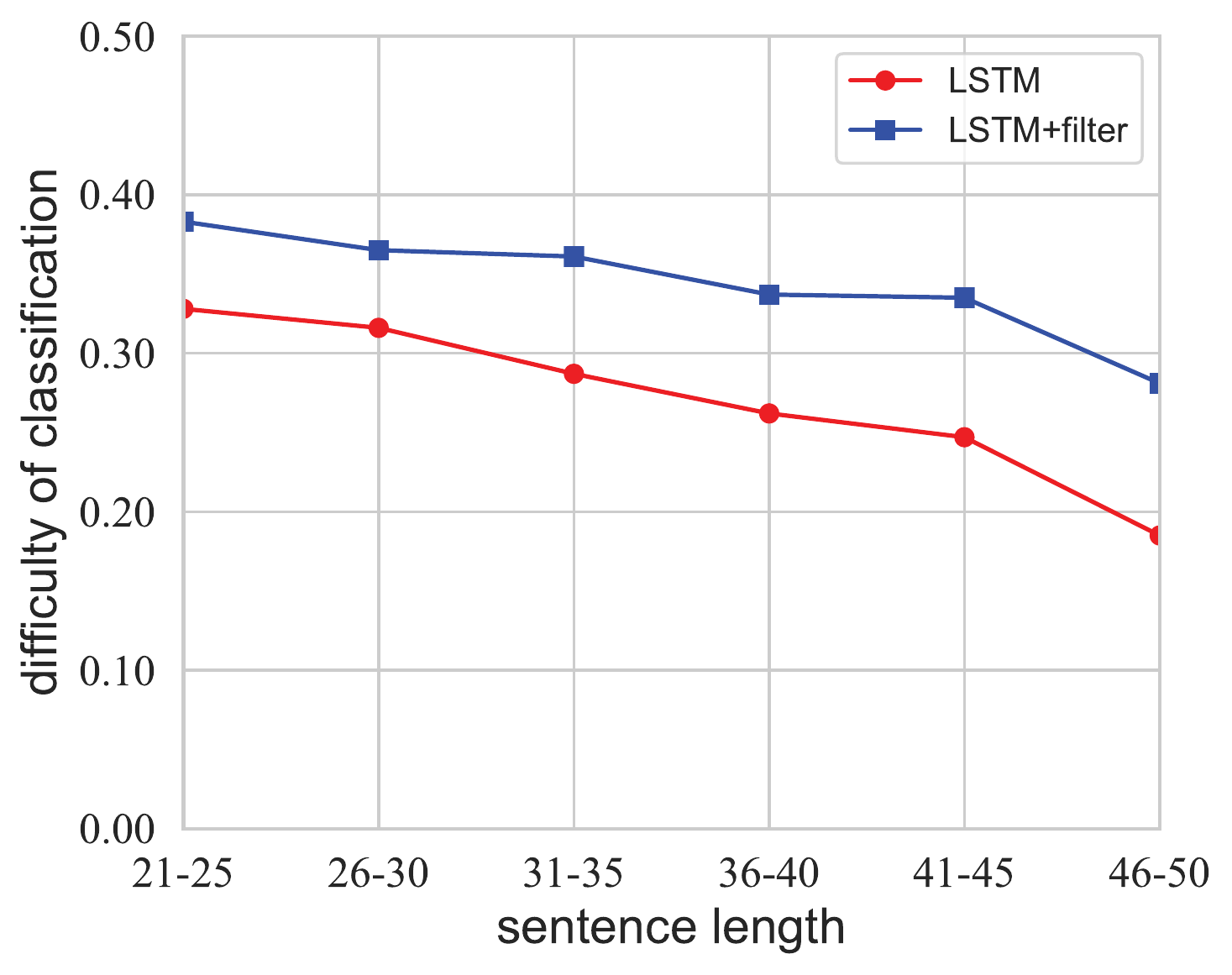}
\caption{The difficulty of classification at various sentence length. The longer the sentence is, the easier the classification is.}
\label{fig:lamda3futu}
\end{minipage}
\hfill
\begin{minipage}[t]{0.48\linewidth}
\centering
\includegraphics[width=0.8\columnwidth]{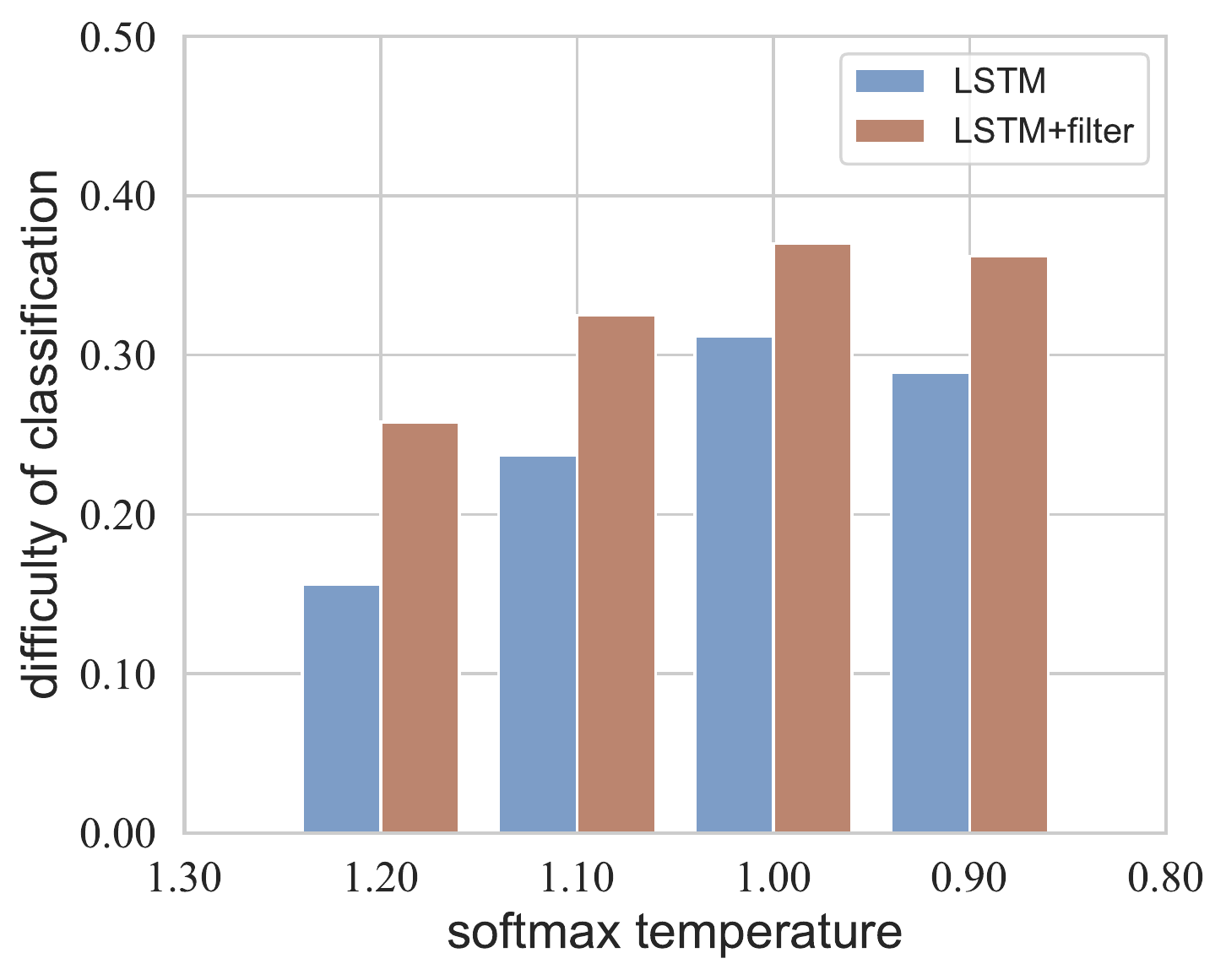}
\caption{The difficulty of classification at various temperature. The further the temperature is from 1.0, the more the difficulty increases.}
\label{fig:acc}
\end{minipage}%
\end{figure}

The difficulty of classification is defined as the classification error rate. The greater the difficulty, the smaller the discrepancy. To observe the change of classification difficulty before and after filtering, we train another classifier with the accepted sentences and real ones. Figure \ref{fig:acc} show that the difficulty increases once the filter is added. %nevertheless of acceptance ratio and softmax temperature. For all these results, LSTM is the generator.

%When temperature is set 1.0, the difficulty increases as the acceptance ratio decreases. The smaller acceptance ratio means the more generated samples will be considered to be rejected. The right two columns of table \ref{t:accept&accuracy} summary the difficulties on two data sets.
Our mechanism works better on longer sentences than the shorter ones. We inspect the difficulty variety as the length of sentences on EMNLP WMT News data set. Figure \ref{fig:lamda3futu} illustrates the results. This indicates that the generation errors accumulate as the sentences become longer. Thus, it is easier for discriminator to make prediction.

Further, we observe the difficulty at different softmax temperatures with the same acceptance ratio. No matter what the temperature is, the classification becomes more difficult after the filtering.  Figure \ref{fig:acc} illustrates the results of LSTM on EMNLP WMT News. In practice, the high temperature is usually used to bias the sample quality. Thus, there is more space for our method to improve the baseline model.
%The classification  error ratio increases 10.2 and 7.3 percent at temperature 1.2 and 0.9 respectively, while it only increases 3.3 percent at temperature 1.0. This explains why the FED score decreases more when the temperature is further from 1.0 in the above section.
%Finally, we find our mechanism works better on longer sentences than the shorter ones.

%The reason is that it is easier for discriminator to predict when more contexts information are encoded.

\section{Related Work}
Autoregressive language models (ALM) have been widely used as text generators for unconditional text generation \cite{Graves2013rnn,Zellers2019}. However, ALM is trained with MLE by being fed the ground truth tokens while no ground truth is available during the reference stage. A series of language GANs are proposed to improve ALM by updating its parameters with a discriminator \cite{Yu2016SeqGAN,Fedus2018MaskGAN,Nie2019ICLR}. The drawback of this way is the reward sparsity and high variance cause mode collapse.

%ALM generates some sentences, a discriminator is trained with these generated sentences and real ones together. Next, this discriminator predicts new generated sentences. These predictions are used to improve ALM and generates better samples. These better samples are mixed with real ones together to train a better discriminator. Repeating this process, until both ALM and discriminator convergence.

Some researchers argue language GANs fail when they are evaluated precisely. Considering  BLEU and self-BLEU only focus on local consistency, \cite{DBLP:journals/corr/abs-1804-07972} designs a language model score and a reverse language model score for evaluating generated text on global semantics. Evaluated by these two paired metrics, all language GANs no longer show any improvement\cite{Semeniuta2018On}. Furthermore, \cite{Caccia2019falling} adjusts the temperature of ALM to trade-off the quality and diversity and find a well-trained language model beats all language GANs. Although \cite{dAutu2019Scratch} trains a language GAN from scratch, its performance is only comparable with ALM. They propose a single metric (FED) and suggest all three metrics should be used together to compare models.

%Inspired by the discriminator can detect the discrepancy between the generated samples and real ones,
A few works improve ALM by using discriminator to filter some generated samples rather than updating the generator's parameters. For text abstraction, \cite{Scialom2020Abstractive} trains this filter to help generator select good token at every generation step with a beam search. \cite{Deng2020EBM} uses the energy-based models to improve the distribution of generated texts. They globally optimize the ALM with an energy term.  A similar work to us is \cite{Caccia2019falling}. They filter the entire sequence by using generator rejection sampling. Regretfully, they compute these samples' likelihood under the generator’s own distribution rather than the discrepancy with the real samples. They also try discriminator rejection sampling. But, this does not always work well except the best one found with a hyper parameter search of 100 trials.  %We do not filter until the whole sequence is generated.

%The reason of this method work well is that their goal biases on the quality in the view of evaluation, otherwise the sample diversity is as important as the sample quality for unconditional text generation.

\section{Discussion}
We create a new generator by adding a filter, which is based on a well-trained discriminator, to a text generator which is an autoaggressive neural language model trained with MLE. This filter rejects some generated samples that have a large discrepancy predicted by discriminator. The accepted samples are the output of this new generator. The distribution of the new generator matches the distribution of real text better than the previous generator. A RNN-based generator and a Transformer-based generator were experimented with two benchmark data sets. The results show these two auotagressive neural language models are improved reliably in both sample quality and sample diversity.
%In contrast to language GANs, which use a discriminator to update the parameters of the generator directly,

There is plenty of scope for future work. The generator may benefit from a more powerful text classifier. A successful case is fake news detection by using a discriminator which has the same architecture as the generator\cite{Zellers2019}. We will try to train the filter and the discriminator in a GAN framework which is suggested in section 2.4. Finally, improving the conditional text generation is worthy of consideration.

\begin{comment}
\section*{Broader Impact}
This paper demonstrates the filter based on the discriminator can effectively improve the unconditional text generator. This provides a very powerful tool for this task.

The methods and ideas proposed in this paper can help scholars and developers studying text generation and help them improve the performance.

The experimental results show that this novel mechanism works reliable across two benchmark data sets by precise evaluation.

Our method does not leverage any bias in the data.
\end{comment}

\bibliographystyle{abbrv}
\bibliography{acl2020}

\begin{thebibliography}{10}

\bibitem{bahdanau2014neural}
D.~Bahdanau, K.~Cho, and Y.~Bengio.
\newblock Neural machine translation by jointly learning to align and
  translate.
\newblock In {\em ICLR}, 2015.

\bibitem{Bengio2015Scheduled}
S.~Bengio, O.~Vinyals, N.~Jaitly, and N.~Shazeer.
\newblock Scheduled sampling for sequence prediction with recurrent neural
  networks.
\newblock In {\em NeurIPS}, 2015.

\bibitem{Caccia2019falling}
M.~Caccia, L.~Caccia, W.~Fedus, H.~Larochelle, J.~Pineau, and L.~Charlin.
\newblock Language gans falling short.
\newblock In {\em ICLR}, 2020.

\bibitem{Chen2015coco}
X.~Chen, H.~Fang, T.-Y. Lin, R.~Vedantam, S.~Gupta, P.~Doll\`ar, and
  L.~Zitnick.
\newblock Microsoft coco captions: Data collection and evaluation server.
\newblock {\em arXiv preprint arXiv:1504.00325}, 2015.

\bibitem{DBLP:journals/corr/abs-1804-07972}
O.~C{\'i}fka, A.~Severyn, E.~Alfonseca, and K.~Filippova.
\newblock Eval all, trust a few, do wrong to none: Comparing sentence
  generation models.
\newblock {\em CoRR}, abs/1804.07972, 2018.

\bibitem{Clark2019Electra}
K.~Clark, M.-T. Luong, Q.~Le, and C.~Manning.
\newblock Electra: Pre-training text encoders as discriminators rather than
  generators.
\newblock In {\em ICLR}, 2019.

\bibitem{dAutu2019Scratch}
C.~d'Autume, M.~Rosca, J.~Rae, and S.~Mohamed.
\newblock Training language gans from scratch.
\newblock In {\em NeurIPS}, 2019.

\bibitem{Deng2020EBM}
Y.~Deng, A.~Bakhtin, M.~Ott, A.~Szlam, and M.~Ranzato.
\newblock Residual energy-based models for text generation.
\newblock In {\em ICLR}, 2020.

\bibitem{Du2019}
W.~Du and A.~Black.
\newblock Boosting dialog response generation.
\newblock In {\em ACL}, page 38–43, 2019.

\bibitem{Fedus2018MaskGAN}
W.~Fedus, I.~Goodfellow, and A.~M. Dai.
\newblock Maskgan: Better text generation via filling in the \_\_\_\_\_\_.
\newblock In {\em ICLR}, 2018.

\bibitem{Ghosh2017affect}
S.~Ghosh, M.~Chollet, E.~Laksana, L.-P. Morency, and S.~Scherer.
\newblock Affect-lm: A neural language model for customizable affective text
  generation.
\newblock In {\em ACL}, pages 634--642, 2017.

\bibitem{Goodfellow2014Generative}
I.~J. Goodfellow, J.~Pouget-Abadie, M.~Mirza, X.~Bing, D.~Warde-Farley,
  S.~Ozair, A.~Courville, and Y.~Bengio.
\newblock Generative adversarial nets.
\newblock In {\em NeurIPS}, 2014.

\bibitem{Graves2013rnn}
A.~Graves.
\newblock Generating sequences with recurrent neural networks.
\newblock {\em arXiv preprint arXiv:1308.0850}, 2013.

\bibitem{He2009Imbalaced}
H.~He and E.~Garcia.
\newblock Learning from imbalanced data.
\newblock {\em IEEE Transactions on Knowledge and Data Engineering},
  21(9):1263--1284, 2009.

\bibitem{hochreiter1997long}
S.~Hochreiter and J.~Schmidhuber.
\newblock Long short-term memory.
\newblock {\em Neural computation}, 9(8):1735--1780, 1997.

\bibitem{Kim2014CNN}
Y.~Kim.
\newblock Convolutional neural networks for sentence classification.
\newblock In {\em EMNLP}, 2014.

\bibitem{Lai2015tc}
S.~Lai, L.~Xu, K.~Liu, and J.~Zhao.
\newblock Recurrent convolutional neural networks for text classification.
\newblock In {\em AAAI}, pages 2267--2273, 2015.

\bibitem{Liu2017Auto}
Y.~Liu, Z.~Qin, W.~Tao, and Z.~Luo.
\newblock Auto-painter: Cartoon image generation from sketch by using
  conditional wasserstein generative adversarial networks.
\newblock {\em Neurocomputing}, 311:78--87, 2018.

\bibitem{Lu2018Past}
S.~Lu, Y.~Zhu, W.~Zhang, J.~Wang, and Y.~Yu.
\newblock Neural text generation: Past, present and beyond.
\newblock {\em arXiv preprint arXiv:1803.07133}, 2018.

\bibitem{mikolov2010recurrent}
T.~Mikolov, M.~Karafi{\'a}t, L.~Burget, J.~{\v{C}}ernock{\`y}, and
  S.~Khudanpur.
\newblock Recurrent neural network based language model.
\newblock In {\em Eleventh annual conference of the international speech
  communication association}, 2010.

\bibitem{Nie2019ICLR}
W.~Nie, N.~Nina, and A.~Patel.
\newblock Relgan: Relational generative adversarial networks for text
  generation.
\newblock In {\em ICLR}, 2019.

\bibitem{Papineni2002BLEU}
K.~Papineni, S.~Roukos, T.~Ward, and W.~J. Zhu.
\newblock Bleu: a method for automatic evaluation of machine translation.
\newblock In {\em ACL}, 2002.

\bibitem{Radford18}
A.~Radford, K.~Narasimhan, T.~Salimans, and I.~Sutskever.
\newblock Improving language understanding by generative pre-training.
\newblock {\em Technical report}, 2018.

\bibitem{Radford19}
A.~Radford, J.~Wu, R.~Child, D.~Luan, D.~Amodei, and I.~Sutskever.
\newblock Language models are unsupervised multitask learners.
\newblock {\em Technical report}, 2019.

\bibitem{Scialom2020Abstractive}
T.~Scialom, P.-A. Dray, S.~Lamprier, B.~Piwowarski, and J.~Staiano.
\newblock Discriminative adversarial search for abstractive summarization.
\newblock {\em arXiv preprint arXiv:2002.10375}, 2020.

\bibitem{Semeniuta2018On}
S.~Semeniuta, A.~Severyn, and S.~Gelly.
\newblock On accurate evaluation of gans for language generation.
\newblock {\em arXiv preprint arXiv:1806.04936}, 2018.

\bibitem{Vaswani2017Attention}
A.~Vaswani, N.~Shazeer, N.~Parmar, J.~Uszkoreit, L.~Jones, A.~N. Gomez,
  L.~Kaiser, and I.~Polosukhin.
\newblock Attention is all you need.
\newblock In {\em NeurIPS}, 2017.

\bibitem{Vinyals2015show}
O.~Vinyals, A.~Toshev, S.~Bengio, and D.~Erhan.
\newblock Show and tell: A neural image caption generator.
\newblock In {\em CVPR}, 2015.

\bibitem{Wu2016Google}
Y.~Wu, M.~Schuster, Z.~Chen, L.~Quoc, N.~Mohammad, M.~Wolfgang, K.~Maxim,
  Y.~Cao, Q.~Gao, and M.~Klaus.
\newblock Google's neural machine translation system: Bridging the gap between
  human and machine translation.
\newblock {\em arXiv preprint arXiv:1609.08144}, 2016.

\bibitem{Yu2016SeqGAN}
L.~Yu, W.~Zhang, J.~Wang, and Y.~Yu.
\newblock Seqgan: Sequence generative adversarial nets with policy gradient.
\newblock In {\em AAAI}, 2017.

\bibitem{Zellers2019}
R.~Zellers, A.~Holtzman, H.~Rashkin, Y.~Bisk, A.~Farhadi, F.~Roesner, and
  Y.~Choi.
\newblock Defending against neural fake news.
\newblock {\em arXiv preprint arXiv:1905.12616}, 2019.

\bibitem{Zhu2018Texygen}
Y.~Zhu, S.~Lu, Z.~Lei, J.~Guo, W.~Zhang, J.~Wang, and Y.~Yong.
\newblock Texygen: A benchmarking platform for text generation models.
\newblock In {\em SIGIR}, 2018.

\end{thebibliography}

\clearpage
\appendix
\section{Results of GPT-2 Evaluated with Local Metric}
%Our model consistently outperforms two neural language models trained with MLE and language GANs on three metrics across two benchmark data sets. More results presented here as appendix.

\begin{figure}[ht]
\centering
\subfigure[COCO Image Captions]{
\includegraphics[width=0.47\columnwidth]{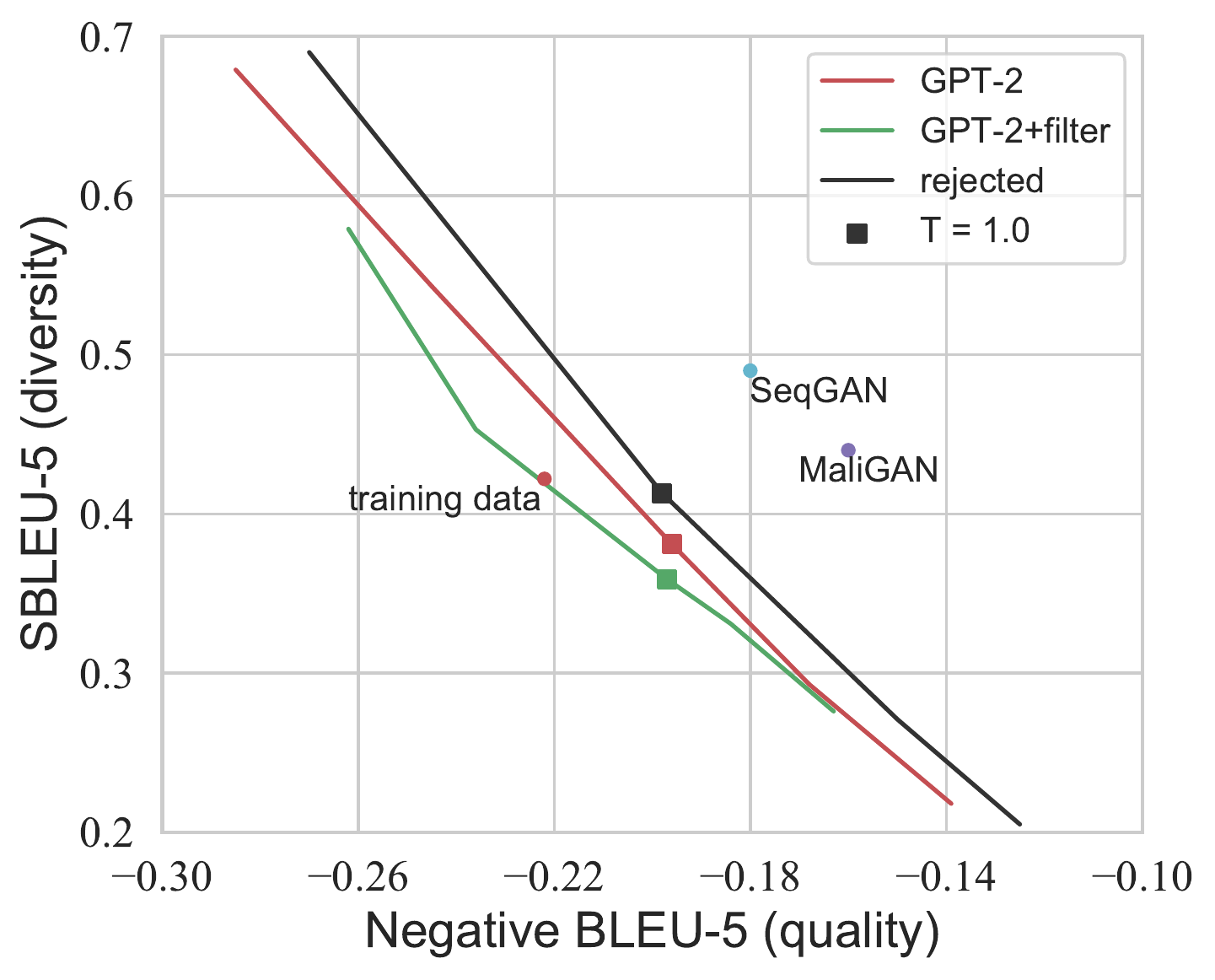}
}
\quad
\subfigure[EMNLP2017 WMT News]{
\includegraphics[width=0.47\columnwidth]{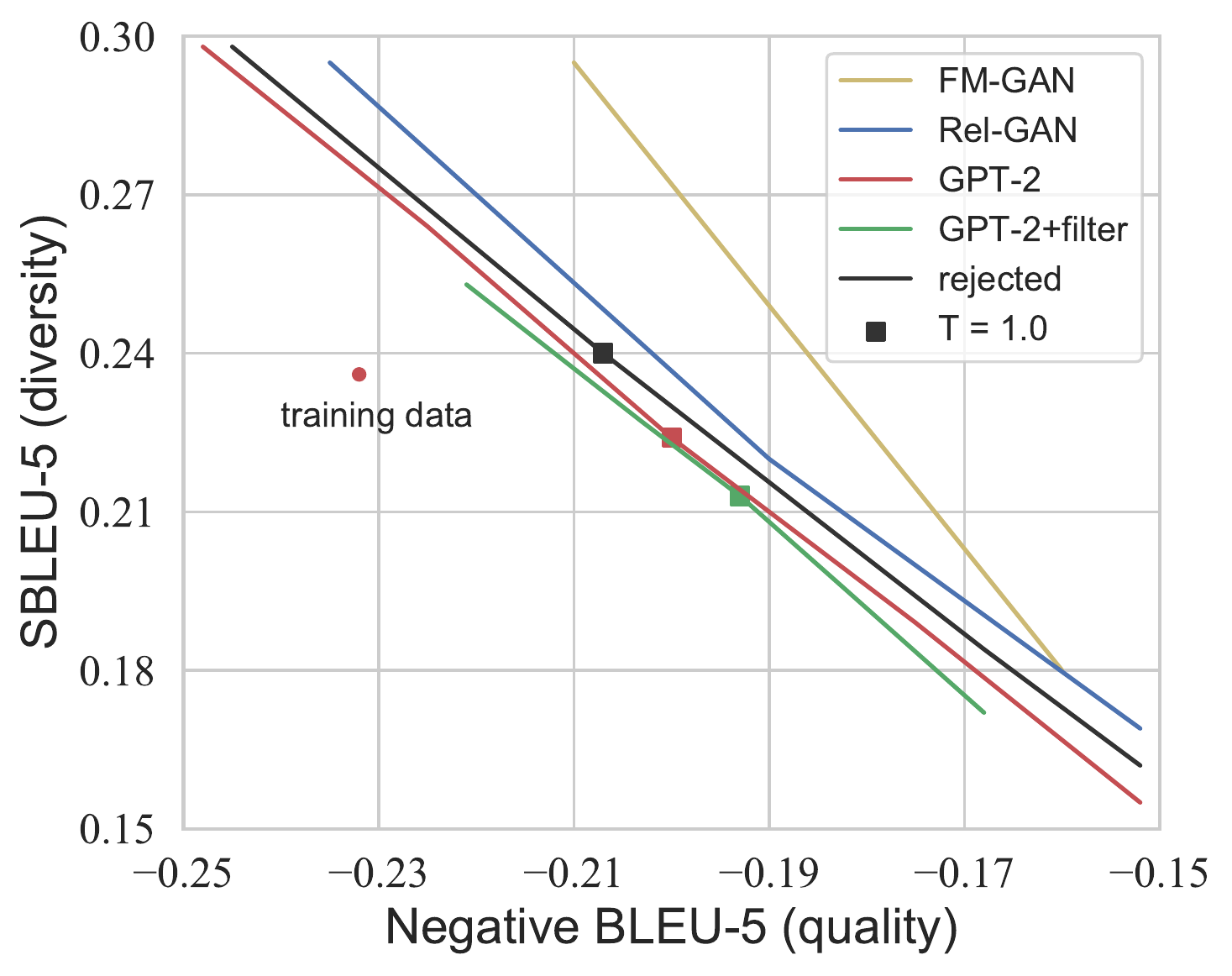}
}
\caption{Local metrics. Lower is better. The results of language GANs in left figure are taken from (Lu et al., 2018); in right figure, they are taken from (Caccia et al., 2020) directly. $T=1.0$ denotes softmax temperature is 1.0. The baseline model is GPT-2.}
\label{figAPP:final}
\end{figure}

For local metric, BLEU versus self-BLEU, we further present the results with GPT-2 as the baseline model across two data sets. Figure~\ref{figAPP:final} shows our model outperforms the baseline model again, let alone those language GANs. It is noted that those GANs fine-tuning LSTM not GPT-2. The performance of rejected samples is plotted as black lines.

\section{Results on COCO Image Captions Evaluated with Global Metric}
\begin{figure}[ht]
\centering
\subfigure[LSTM]{
\includegraphics[width=0.47\columnwidth]{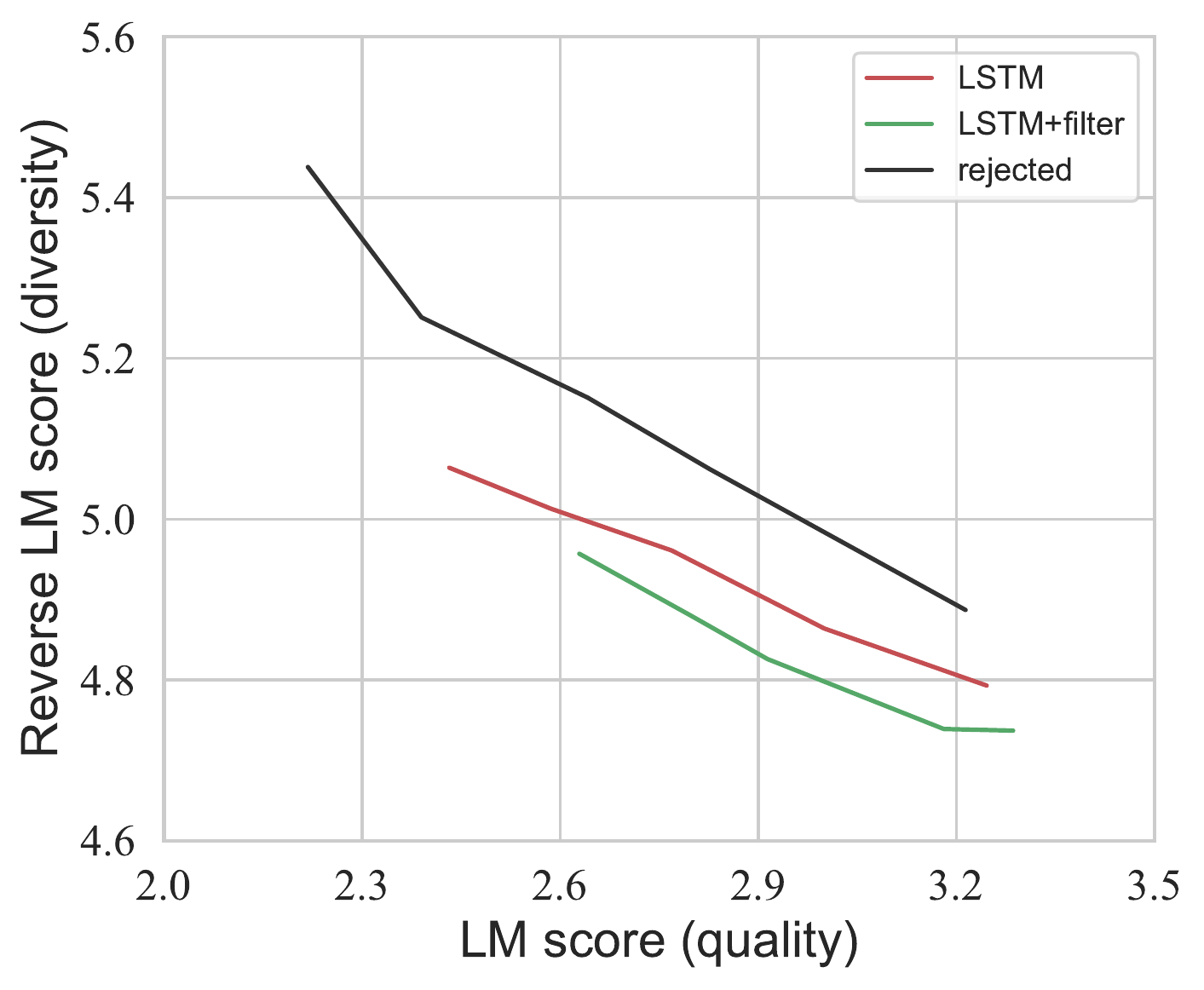}
}
\quad
\subfigure[GPT-2]{
\includegraphics[width=0.47\columnwidth]{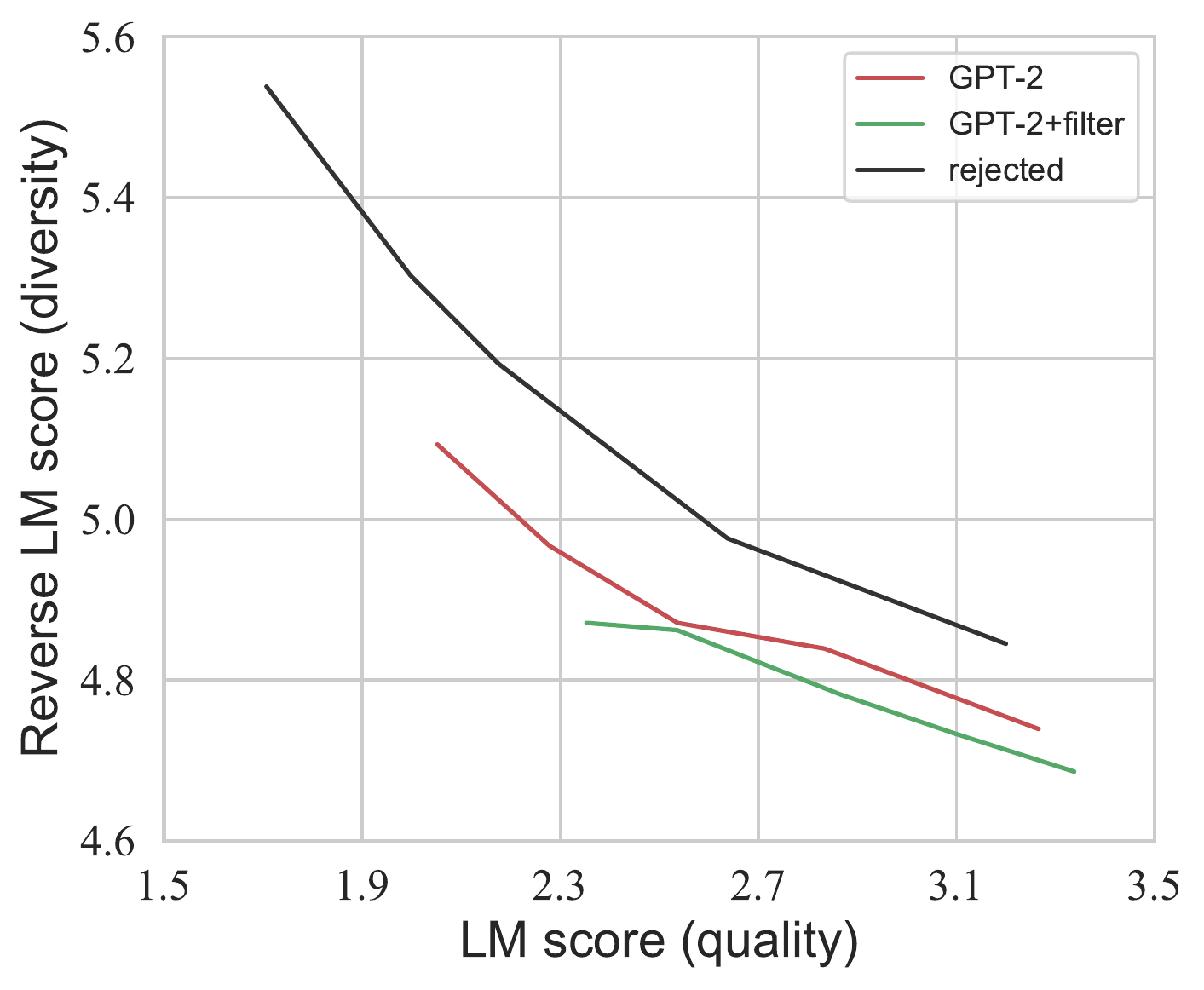}
}
\caption{Global metrics. Lower is better. The data set is COCO Image Captions.}
\label{figAPP:LMRLM}
\end{figure}

For global metric, language model score versus reverse language model score,  we further present the results on COCO Image Captions with both GPT-2 and LSTM as the baseline models. Figure \ref{figAPP:LMRLM} illustrates the results. New generators always outperform the old ones and language GANs. Due to the limitation of space, we do not present the results of temperature less than 1.0. The performance of rejected samples is plotted as black lines.

%The results on COCO Image Captions are supplemented. %和EMNLP相比，效果更明显？？

\section{More Results About The Influence of Acceptance Ratio }
The filtering mechanism always works well regardless of the value of acceptance ratio. Acceptance ratio plays a role in balancing quality and diversity, and has a huge impact on them. Thus, this provides us a new way to adjust the sample quality and sample diversity. Acceptance ratio can be used in conjunction with the temperature to obtain better generation performance.
%The acceptance ratio, i.e. the number of the accepted sentences divided by the total generated sentences is efficiency of filtering mechanism. Obviously, the efficiency becomes low as $c$ decreases.

%Figure~\ref{fig:bsb} shows the influence of the acceptance ratio with LSTM on the EMNLP WMT News data set. %Figure~\ref{fig:bsb} and Figure~\ref{fig:bsb} show the results of GPT-2 on COCO Image Captions and EMNLP WMT News respectively.

%We find all points move towards performance point of the training data. %Our model even achieves the BLEU score verse self-BLEU score almost equal to scores of the training data at temperature 1.2. This highlights the effectiveness of our method.
% % Before that, only what we can do is adjusting the softmax temperature.
%On the contrary,  the quality is improved greatly and the diversity is sacrificed a little, when the temperature is less than 1.0.

\begin{figure}[!b]
\centering
\subfigure[LSTM on EMNLP2017 WMT News]{
\includegraphics[width=0.48\columnwidth]{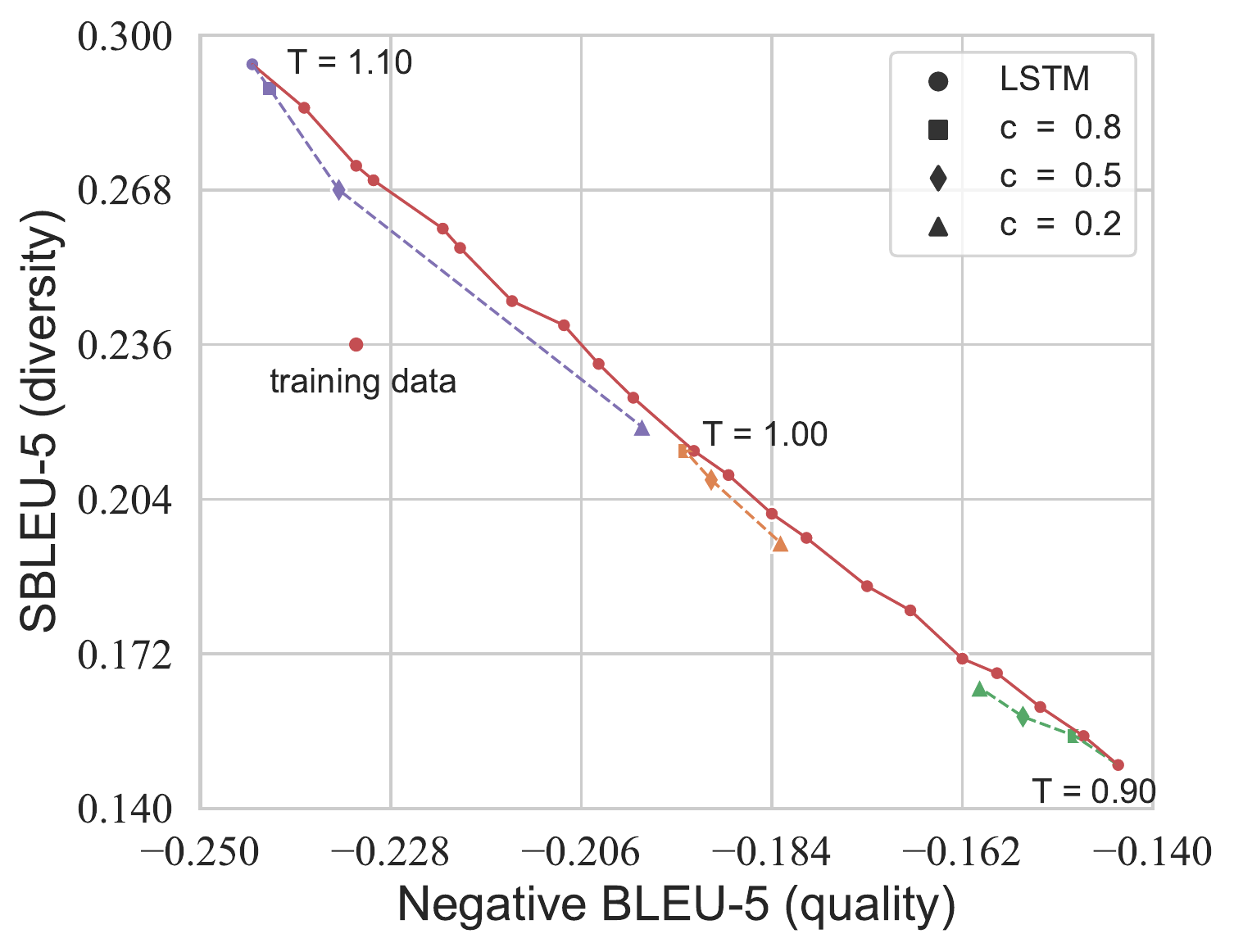}
}
\quad
\subfigure[GPT-2 on COCO Image Captions]{
\includegraphics[width=0.48\columnwidth]{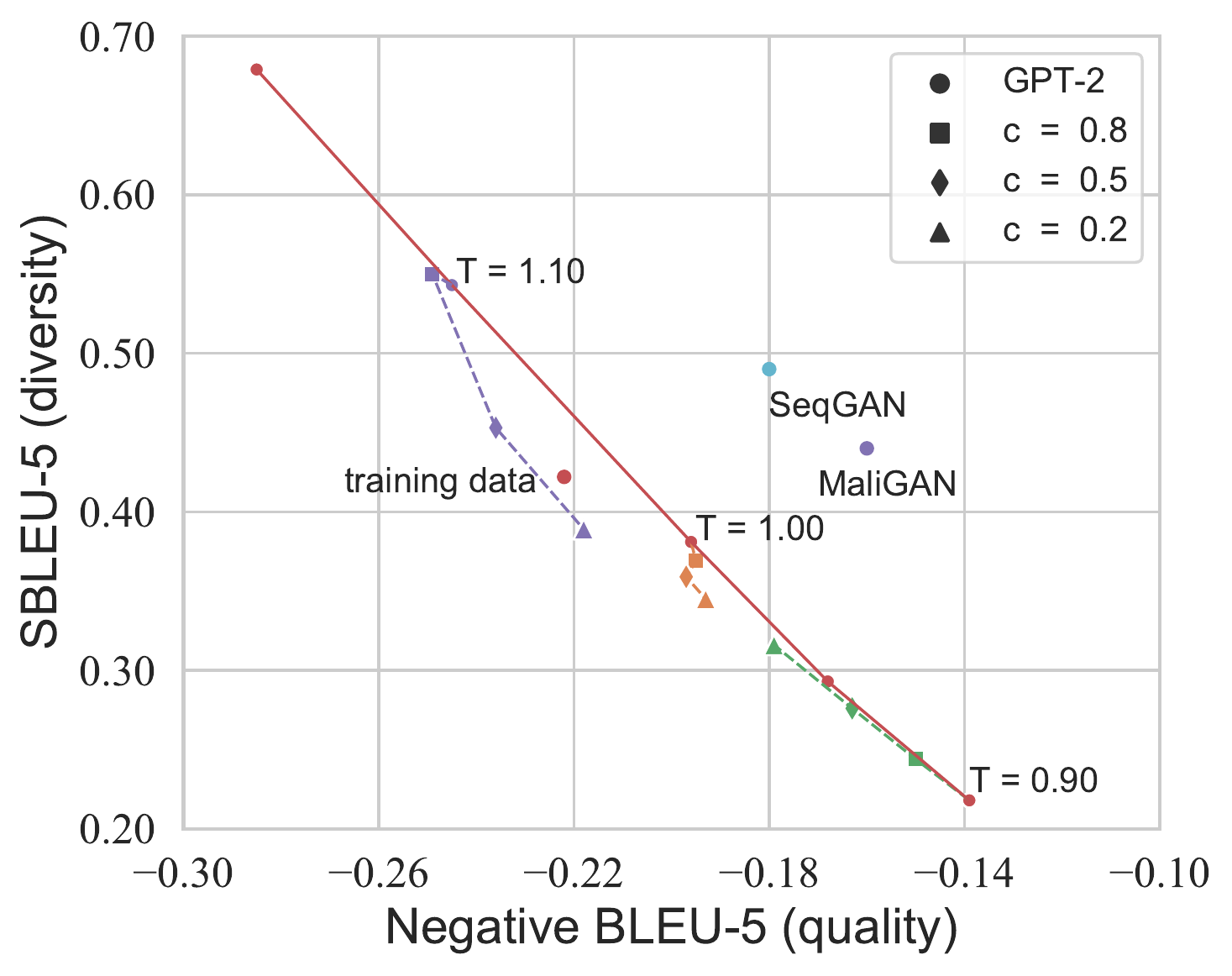}
}
\quad
\subfigure[GPT-2 on EMNLP2017 WMT News]{
\includegraphics[width=0.48\columnwidth]{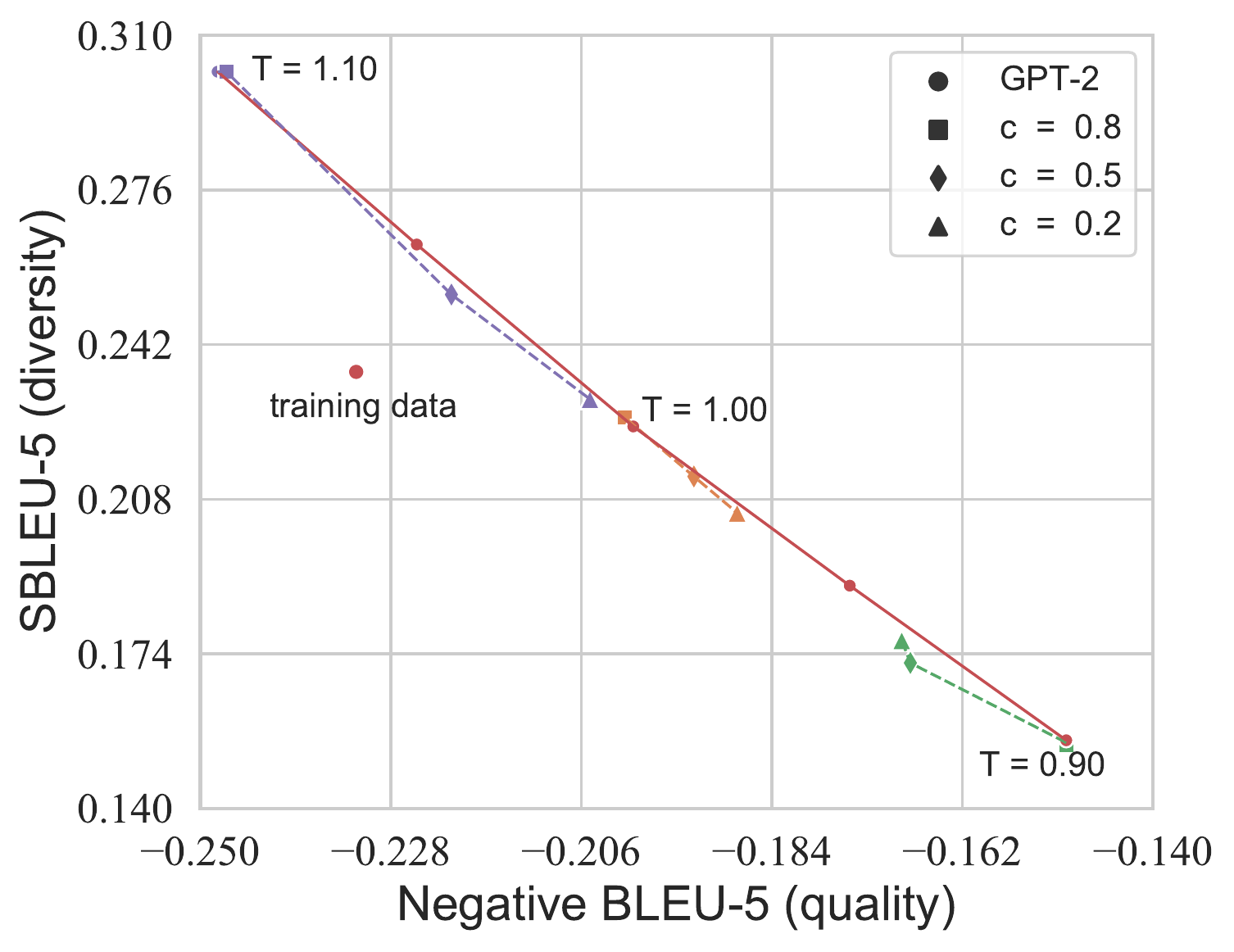}
}
\caption{The Influence of Acceptance Ratio. The results of SeqGAN and MaliGAN are taken from (Lu et al., 2018). }%The results of FmGAN and RelGAN are taken from (Cacci et al., 2020).}
\label{figAPP:bsb}
\end{figure}

\section{EMNLP WMT News Samples}
We list some sentences for EMNLP WMT News.
\begin{table}[h]
    \centering
    \begin{tabular}{p{1cm}p{12cm}}
    %\toprule
          %EMNLP News \\
         \hline
         \hline
         LSTM &  He is by reaction , cannot say about one track union ; what can you win ? never him simply a few words .\\
         & The children of Mr Taylor ' s parents were also injured because of an enormous 20 weeks of professional abuse in England .\\
         & She came back , said Don ' t worry about all your problems on the impact , she noted in the 2015 weeks .\\
         & 31 UPI Well , until the cold weather , something was on display go and so I have to talk .\\
         & I believe in a short period of time to say the only thing that happened over that years was that we want them to be able to get together and defend them .\\
         & He stress that the former president elect won a senator with an opinion of Monday ' s father of his public .\\
         & I ' ve been speaking out , and in my digital life , it was done in the last 12 months .\\
         & And the bank says that the money for Amazon can help the economy around the world , as well as revenue of those jobs .\\
         & These types of games have contributed to the poor transition of arms watch that they bought as their asset gains and solid .\\
         & The National Park Service had profit of 2 . 6 million , a small proportion of the population , encouraged or that the Chinese had been cut up from the factory .\\
         & She said they had come to realise that she was still hiding off the door , but it was even fallen .\\
         & While Jordan remains popular for a generation , he ' s been in odds with top three women now since their year .\\
         & It ' ll be a different assessment of players and international coaches if they ' re the right ratings , Mourinho said .\\
         & Facebook ' s executive chief announced on Saturday that it would soon take up to 50 years to pass controversy in America this week .\\
         & In interviews given Trump ' s domestic war family , he said In the next man , the big path on our road is too high .\\
         & But I couldn ' t talk about answers to him and you hear about the thing of you , and and I think it was unlikely he would have read ? \\
         & It doesn ' t think that we have such a strange future for people this year , about how we look at the Welsh Government . \\
         & People will seek to work to write in a side of the same question , and they ' ve got a good idea of those decisions , he said .\\
         & If you do not talk about how violent victims are in risk of an assault made online , we will be working with anyone else to do it .\\
         & We have been asked him to match him in two games with the face of Twitter and I guess he speaks of a question .\\
         & ' I already don ' t care but a world and if you don ' t have good luck , you can see everything from my peers .\\
         & We are confident that Israel will achieve its best level of relations with Israel and continue to consider their Syria relationships , the Economic profile of the US said .\\
         & This week , on July 21 , he offered 100 percent of his money in the two years to second quarter .\\
         & Danny Baker says he should have played some learning and he would remember that or letting her back on 5 , 000 and 0 1 .\\
         & It was quickly sent back on June 11 over 24 billion on the annual original initiative , which the foundation introduced at the end of those years .\\
    \bottomrule
    \end{tabular}
    \caption{Samples from LSTM trained with MLE}
    \label{tab:my_label}
\end{table}

\begin{table}[h]
    \centering
    \begin{tabular}{p{1cm}p{12cm}}
    %\toprule
          %EMNLP News \\
         \hline
         \hline
         LSTM + filter &  I had little doubt that these states should make the best possible self worth than those who pay for the costs of net debt .\\
         & It maintained , which is worth ten billion in output at 42 million in buyers , with huge growth slightly from 1 . 8 billion in March .\\
         & As he put on a DNA studio in an interview , he said the pressure had a lot from learning and outside .\\
         & The gains in oil tax growth increase from more than 0 . 2 percent in the weeks after the Brexit vote .\\
         & The club experienced attempted criticism of the injury after winning five penalties again in England without coming point after losing the game .\\
         & Since polls have been carried out , the 10 year old could have donated 21 billion from her fund for the government .\\
         & For full time the exhibition has more than 600 homeless residents who died in crimes and deadly police and local community officials in New Jersey .\\
         & Those are Australia ' s most powerful communities , well below the average level of UK population , and this was only half , and it was on the lead .\\
         & It ' s sad that they haven ' t looked at their fate or won ' t happen on for several years .\\
         & The programme number one was not to get in touch with any political foreign minister , Prime Minister Theresa May told the Commons ' s Home Office .\\
         & So then I hope to get tough under Leicester ' s belt and make it very difficult for the players to accept that much more week and day on .\\
         & Trump has repeatedly stepped under pressure calls for a terrorist attack , but could also take people in the Oval Office to update the news .\\
         & These agree , providing support to Hollande and the union might not mean it for America , or is nice to create to embrace the strongest of the world ' s biggest trading partners .\\
         & They would hold them higher at 2 . 2 in the July holiday cycle when they could stay on the line to pay again for comfort in 2016 .\\
         & The Spanish authorities are in control of civilian areas throughout the spring , but took away another group of tourists .\\
         & The February 12 vote also because he had been through months in early voting was attended by all of us .\\
         & Police arrested the suspect and said his body was banned and was being treated as he might face serious injuries when he was released .\\
         & One shares of the year said the company could open the annual stock market with a record of 2 . 4 billion .\\
         & The blast to the man on Tuesday night was captured during the day , and it wasn ' t revealed whether any information or arrests were made .\\
         & I don ' t want to give people an option but whether I ' m re elected , it has a lot of vision for Britain .\\
         & Some 74 , 000 people have had there via the debt from 2013 in the past year alone , according to a new report .\\
         & If you ' ve found joy , we have been the best job to be left behind them , and we decided to talk about very much when anything so different or cool .\\
         & And to see their favourite community ends and get a lot of these injured if they win , they have to continue that .\\
         %& Officials say they have won a budget shooting , which will examine what digital information is for them and sell in a safe place for each other .\\
    \bottomrule
    \end{tabular}
    \caption{Samples from LSTM+filter}
    \label{tab:emnlp_accepte}
\end{table}

\begin{table}[ht]
    \centering
    \begin{tabular}{p{1cm}p{12cm}}
    %\toprule
          %EMNLP News \\
         \hline
         \hline
         Rejected &  A result of some million people and any man is in Brazil , US and the largest types of water and sea .\\
         & As well as a doctor she described the as before the results The holes came or the false support was not the same .\\
         & In 50 years , in the last two years , he ' s outstanding to me , credit and remember in New York .\\
         & I have to see this color telling you I am familiar just when you feel like you ' re going after great things .\\
         & If it had advanced in the space industry , suddenly it wasn ' t going to let people say getting orders from Christmas ' s sheet was the S P 5 . . .\\
         & I was happy to have settled writing fairly hard letters .\\
         & Trump broke all of our last time , in order for the U . S . led coalition to honor Russia that are very angry about that .\\
         & We recognise the more fear that all the people deserve from the branch but that they can benefit everything to make them feel safe .\\
         & Since I ' ve shown that it has been impossible knowing in the last few days , whether the greatest goal is in it before .\\
         & The airport has no intention of being installed and a failure of the international property is actually expected to return to the future for the first time as well .\\
         & At the beginning of the day , if you want to live in a tiny town , you have to name a different cup , he said .\\
         & My daughter gives birth to work at such a critical point ; she will not wish to be something she calls for any UK pay .\\
         & The sixth highest and 300 is reported to increased steady revenue , and was down from 66 . 7 million this year 18 . 9 percent last year .\\
         & So here ' s what I might be able to see and I talk to Boris , he is an iconic pro active president .\\
         & If you ' re outside my hotel and then call him you all go about the clock towards a protest in the stands at the park and kill him , he said .\\
         & Despite the fact that she has a free trade deal but that she does the right thing , at the moment she ' s going to and stronger up to remain .\\
         & There is a lot of actors , particularly around the world who have enough freedom of understanding over the next year , she told reporters in speech .\\
         & He said he was very excited and ready about how the Brumbies would get him in a third half in the league and the numbers from early Monday .\\
         & Many of both sides of those things are trying to record out of that or generally about Trump and his personal family .\\
         & This year ' s media story is likely to help focus on most of the film ' s artists who just will easily realise how much she used to encourage her to share pain at some stage in all .\\
         & Anna knew that we opened a hotel with wonderful weapons in the library , which has seen them by free for an event left mostly fighting .\\
         & This is great to happen if we have a good terms of health for the need and good aid is to the point of them .\\
    \bottomrule
    \end{tabular}
    \caption{Samples from Rejected}
    \label{tab:emnlp_rejected}
\end{table}
\clearpage

\section{COCO Image Captions Samples}
We list some sentences for COCO Image Captions.

\begin{table}[h]
    \centering
    \begin{tabular}{p{1cm}p{12cm}}
    %\toprule
          %EMNLP News \\
         \hline
         \hline
         LSTM &  a bathroom with a sink and a shower and a sink and cleaner .\\
         & a kitchen filled with pots and sinks and bamboo walls .\\
         & a mirror with three and large mirrors on it and a gas hole on top of it .\\
         & a public white cat sitting next to each other .\\
         & a bath room with a tiled walls and red checkered rug .\\
         & jumbo police up red to look and pots and picnic above the blank blue bottles .\\
         & a large building with others on it and trees at the wall .\\
         & a kitchen filled with lots cabinet inside to a tile bathroom .\\
         & a large pan with the black comes walking for it .\\
         & a black laying on a southwest plane is walking down .\\
         & american stuffed bikes facing very materials on the rain .\\
         & a bathroom with graffiti and a giant floor top .\\
         & a red toilet on the of white walls and wood remodeled .\\
         & men with looking states bears sitting on a table .\\
         & three open people are preparing four dirt food watching customers .\\
         & a dog is sitting on top of a car .\\
         & a bath and black surfer walking on top of a person .\\
         & three birds parked on a parked snowy sign of .\\
         & a woman getting a bicycle in the middle of a store .\\
         & a bathroom has a toilet , and double sinks .\\
         & a very toilet is surprised next in the process of a bathroom .\\
         & a couch with horserace kickstand and book .\\
         & two extremely dogs next to a window near a small toilet .\\
         & a bathroom with a toilet and a toilet and toilet .\\
         & a plane is flying across top of the the blue with red rims .\\
         & a man and a girl standing next to his bike on a street .\\
         & a group of men grazing on a downhill grass .\\
         & a large metal jet lights at a cloudy sky .\\
         & a bike parked on a counter top of the shade .\\
         & a long framed brown plate topped with a view of the center of a table .\\
         & group of motorcycles parked in a restroom near a church .\\
         & a woman riding a scooter in front of a rural motorcycle .\\
         & a kitchen with two sinks with clutter , pans and window in it .\\
         & a bowl filled with cups parked holds tables and a strawberry .\\
         & stop bathroom with an island shining in a man's fan plan .\\
         & several white birds hanging together , clean trees .\\
         & a person riding a motorcycle at an grass .\\
         & a man looks with a drink in the bathroom .\\
         & a black car parked next to a red bench .\\
         %& a living room with people containers and a white toilet .\\
         %& a bathroom features a fridge and the ceiling comes is all and appliances .\\
         %& motor motor sold and pots and cheese in a large airplane .\\
         %& many airplanes and its war luggage on the hangar .\\
         & a giraffe fixing off near a handicapped sky .\\
         & a plane is standing down no landing in a blue field .\\
         & the interior of kitchen talking in shower next to it's socket and .\\
         & a person standing next to a bike on a street .\\
         & a futuristic benches is along the runway covered .\\
         & three people riding motorcycles sitting in an air pasture .\\
         & a bathroom with a glass mirror with a specially .\\
         & a small hotel room with a carpet , filled in shelves .\\
    \bottomrule
    \end{tabular}
    \caption{Samples from LSTM trained with MLE}
    \label{tab:my_label_COCO}
\end{table}

\begin{table}[h]
    \centering
    \begin{tabular}{p{1cm}p{12cm}}
    %\toprule
          %EMNLP News \\
         \hline
         \hline
         LSTM + filter &  a man sitting inside a road next to a motorcycle .\\
         & a bathroom with mirror to lights and an messy dirty toilet .\\
         & a motorcycle parked on the street in the race .\\
         & a silver sofa sitting under a kitchen counter .\\
         & this is an image of a bathroom with a bright black toothbrush .\\
         & a woman rides a bike down the street .\\
         & a man and black motorcycle and a pink umbrella .\\
         & a living room with two different a leds frame .\\
         & a group of people in motorcycle and one child near a park .\\
         & a women standing next to a kitchen .\\
         & a man in a kitchen looking at a bar with a flower near the store .\\
         & an open restroom with a toilet paper walls and an open shower .\\
         & a blue motorcycle parked next to the side of it of a fence .\\
         & benches on the side of an air port run with grazing .\\
         & white painted clean toilet stall with a bathtub , seats in it .\\
         & there is an older couple of motorcycles parked on the street .\\
         & an overheard view of a kitchen that includes a hot cupboard , a stove .\\
         & a picture of a silver toilet that is installed .\\
         & a bathroom with a sink and a can of milk .\\
         & a man standing next to each other by a cat .\\
         & a little girl in a kitchen looking at the plants and lotions .\\
         & a man in a farmers shirt looking all sitting on the beach .\\
         & there are a dog sitting on the wheel of a crowd of the bicycle .\\
         & a cop sitting holding a horse on the trailer overlooking the water .\\
         & a person walking across the street in a v there of a poor street .\\
         & a man and a dog look over a boat .\\
         & a small bath room with white cabinets is and basin grate and green countertop range and storage walls .\\
         & a narrow kitchen with lots of pots and pans .\\
         & a photo of a man riding on a commercial motorcycle .\\
         & a bathroom with a pink bathtub and the toilet . \\
         & a woman sitting at the curb in a toilet .\\
         & a black motorcyclist that is on a motor bike .\\
         & an orange cat sitting on a bench on a toilet .\\
         & a person resting on a motor bike across the road .\\
         & a small black helicopter on an open field .\\
         & an airplane is flying high in a clear sky .\\
         & a kitchen with recessed flooring dryin furnished , sink and living cabinets .\\
         & two police motorcycles parked on top of a street .\\
    \bottomrule
    \end{tabular}
    \caption{Samples from LSTM+filter}
    \label{tab:coco_accepte}
\end{table}

\begin{table}[h]
    \centering
    \begin{tabular}{p{1cm}p{12cm}}
    %\toprule
          %EMNLP News \\
         \hline
         \hline
         Rejected &  a light desk with a red hot sink and dishwasher .\\
         & this are yellow toilet in a bathroom , an toilet paper empty .\\
         & a white cat chillin sitting in the wall next to a black place .\\
         & a bathroom with a blue refrigerator , a sink and dishwasher over the .\\
         & a large banana sitting near a blue screen field .\\
         & several women riding by motorcycles out of a car .\\
         & a bathroom with a sink , red wash walls .\\
         & a view of a fashioned photograph of a church with two almost use overpass it .\\
         & a motor marble jet moved away from a blue road .\\
         & a dog and black picture of a big airplane and people on a field .\\
         & food , and a woman is holding a brown motorcycle .\\
         & a hotel room sitting with a toilet , sink , a large toilet .\\
         & a white and white colored image of a man attached to a phone .\\
         & a woman wearing whom walking in a public kitchen .\\
         & a clean kitchen filled with a wooden dish and a bottle of cupboards .\\
         & a bathroom is has a metal sink , and white control window .\\
         & two urinals next to some children and umbrellas .\\
         & a woman and man on a motorcycle on the road .\\
         & an image of a person is surfing a banana .\\
         & a rustic kitchen with appliances and oven for windows .\\
         & a little man posing holding bikes in front of a toilet .\\
         & a plane flying through a group of pigtails on the wheel of a silly foot street .\\
         & two work in a public kitchen filled with things .\\
         & several men riding motorcycles are in a grassy field .\\
         & a woman perched on a bench with a white bike and with fuzzy team .\\
         & a green toilet sitting in a bathroom next to a single toilet in a tiny wall .\\
         & a commercial airplane flying high at the air clouds .\\
         & a man and his blue bear is standing outside on a small blue tub .\\
         & a person on a street on the with a toilet .\\
         & a corner of a hotel on the water .\\
         & a downtown room with a stove with toilets in a well cabinets on a restroom .\\
         & a subway bathtuband cowboy down equipment and watch one in the sky .\\
         & a bathroom with yellow cabinets and a gas color in the background .\\
         & a kitchen with a white toilet and modern logo on the wall .\\
         & the black toilet parked side next to a sink and sink .\\
         & a motor cat sitting on top of a large rubber closed block .\\
    \bottomrule
    \end{tabular}
    \caption{Samples from Rejected}
    \label{tab:COCO_rejected}
\end{table}

\end{document}